\documentclass[conference]{IEEEtran}
\IEEEoverridecommandlockouts

\usepackage{bbm}
\usepackage{graphicx}

\usepackage{xcolor}
\usepackage{float}   
\usepackage[normalem]{ulem}
\usepackage{dsfont}
\usepackage{textcomp}
\usepackage{stfloats}
\usepackage{url}
\usepackage{verbatim}
\usepackage{algorithmic}
\usepackage{array}
\usepackage{booktabs}
\usepackage{pifont}
\usepackage{comment}
\usepackage{algorithm}
\usepackage{algorithmic}
\usepackage[acronym]{glossaries}
\newacronym{adc}{ADC}{Analog-to-digital converter}
\newacronym{anns}{ANNs}{artificial neural networks}
\newacronym{bai}{BAI}{best arm identification}
\newacronym{bs}{BS}{base station}
\newacronym{bsc}{BSC}{binary symmetric channel}
\newacronym{bptt}{BPTT}{backpropagation-through-time}
\newacronym{agv}{AGV}{automatic guided vehicle}
\newacronym{cdma}{CDMA}{code division multiple access}
\newacronym{cdf}{CDF}{cumulative distribution function}
\newacronym{comos}{CMOS}{complementary metal-oxide-semiconductor}
\newacronym{csnn}{CSNN}{convolutional spiking neural networks}
\newacronym{cnn}{CNN}{convolutional neural network}
\newacronym{dvs}{DVS}{dynamic vision sensor}
\newacronym{dl}{DL}{downlink}
\newacronym{ea}{EA}{estimated anomaly-only}
\newacronym{es}{ES}{edge server}
\newacronym{fl}{FL}{federated learning}
\newacronym{flops}{FLOPs}{floating-point operations}
\newacronym{fdr}{FDR}{false discovery rate}
\newacronym{fov}{FOV}{field of view}
\newacronym{fs}{FS}{full-sequence}
\newacronym{gmm}{GMM}{gaussian mixture model}

\newacronym{kl}{KL}{Kullback–Leibler}

\newacronym{mab}{MAB}{multi-armed bandit}
\newacronym{mcmc}{MCMC}{Markov Chain Monte Carlo}
\newacronym{npu}{NPU}{neuromorphic processing unit}
\newacronym{neurosn}{neuro-SN}{neuromorphic sensor nodes}
\newacronym{neurocomm}{NeruoComm}{neuromorphic communication}
\newacronym{ppm}{PPM}{pulse-position modulation}
\newacronym{pdf}{PDF}{probability density function}
\newacronym{pmf}{PMF}{probability mass function}
\newacronym{iid}{i.i.d.}{independent and identically distributed}
\newacronym{iot}{IoT}{internet of things}
\newacronym{lif}{LIF}{leaky integrate-and-fire}
\newacronym{ir}{IR}{impulse radio}

\newacronym{ook}{OOK}{on-off Keying}
\newacronym{snn}{SNN}{spiking neural network}
\newacronym{tdr}{TDR}{true discovery rate}
\newacronym{ul}{UL}{uplink}
\newacronym{wsns}{WSNs}{Wireless Sensor Networks}
\newacronym{spr}{SRM}{Spike Response Model}
\newacronym{tdma}{TDMA}{time division multiple access}
\newacronym{ts}{TS}{Thompson Sampling}

\newacronym{rf}{RF}{Radio Frequency}
\newacronym{rr}{RR-scheduling}{round robin scheduling}

\newacronym{pac}{PAC}{Probably Approximately Correct}
\newacronym{ppp}{PPP}{poisson point process}

\newacronym{ucb}{UCB}{upper confidence bound}
\usepackage{amsmath,amsfonts, amssymb}
\usepackage{bm}
\usepackage{graphicx}

\usepackage[font=small]{caption}
\usepackage[font=footnotesize]{subcaption}
\usepackage{tikz}
\usetikzlibrary{plotmarks,patterns,decorations.pathreplacing,backgrounds,calc,arrows,arrows.meta,spy,matrix}
\usepackage{tikzscale}

\usepackage{pgfplots}
\usepgfplotslibrary{colorbrewer, groupplots, patchplots}
\pgfplotsset{
    compat=newest,
    legend style={font=\footnotesize, fill opacity=0.7,  draw opacity=1, text opacity=1, draw=white!15!black, legend cell align=left, align=left}, 
    width=0.8\columnwidth, 
    scale only axis,
    height=4cm,
    yminorticks=false,
    xminorticks=false,
    label style={font=\small},
    title style={font=\small},
    tick align=outside,
    tick pos=left,
    tick style={color=black},
    tick label style={font=\footnotesize},
    grid style={line width=.1pt, draw=gray!20},
    major grid style={line width=.1pt,draw=gray!20},
    plot coordinates/math parser=false 
}
\newlength\figureheight
\newlength\figurewidth

\usepackage{multirow}
\usepackage{tablefootnote}
\usepackage{booktabs}
\usepackage{tabularx}

\def\BibTeX{{\rm B\kern-.05em{\sc i\kern-.025em b}\kern-.08em T\kern-.1667em\lower.7ex\hbox{E}\kern-.125emX}}

\newcommand{\mb}[1]{\mathbf{#1}}    

\newcommand{\argmin}{\mathop{\rm arg~min}\limits}
\newcommand{\argmax}{\mathop{\rm arg~max}\limits}
\newcommand{\probP}{\text{I\kern-0.15em P}}

\usepackage{amsthm}
\newtheoremstyle{noparens}%
  {}{}
  {\itshape}
  {}
  {}
  {.}
  { }
  {\textbf{\thmname{#1}}\textbf{\thmnumber{ #2}}\thmnote{ #3}}

\theoremstyle{noparens}

\newtheorem{proposition}{Proposition}

\definecolor{amaranth}{rgb}{0.9, 0.17, 0.31}
\begin{document}

\title{Online Reliable Anomaly Detection via Neuromorphic Sensing and Communications}
\author{Junya Shiraishi,~\IEEEmembership{Member,~IEEE,}~Jiechen Chen,~\IEEEmembership{Member,~IEEE,} Osvaldo Simeone,~\IEEEmembership{Fellow,~IEEE,} \\and Petar Popovski,~\IEEEmembership{Fellow,~IEEE}
\thanks{J. Shiraishi and J. Chen contributed equally to this work. The work of J. Shiraishi was supported by European Union's Horizon Europe research and innovation funding programme under Marie Sk{\l}odowska-Curie Action (MSCA) Postdoctoral Fellowship, ``NEUTRINAI" with grant agreement No.~101151067. The work of J. Chen and O. Simeone was supported by the EPSRC project EP/X011852/1, and O. Simeone was also supported by an Open Fellowships of the EPSRC (EP/W024101/1). The work of P. Popovski was supported by the Velux Foundation, Denmark, through the Villum Investigator Grant WATER, nr. 37793. \emph{(Corresponding author: Jiechen Chen)}}
\thanks{J. Shiraishi and P. Popovski are with the Department of Electronic Systems, Aalborg University, 9220 Aalborg, Denmark (e-mail: \{jush, petarp\}@es.aau.dk). J. Chen and O. Simeone are with the King's Communications, Learning and Information Processing Laboratory, King's College London, WC2R 2LS London, U.K. (e-mail: \{jiechen.chen, osvaldo.simeone\}@kcl.ac.uk)}
}

\maketitle

\begin{abstract}
This paper proposes a low-power online anomaly detection framework based on neuromorphic wireless sensor networks, encompassing possible use cases such as brain–machine interfaces and remote environmental monitoring. In the considered system, a central reader node actively queries a subset of neuromorphic sensor nodes (neuro-SNs)  at each time frame. The neuromorphic sensors are event-driven, producing spikes in correspondence to relevant changes in the monitored system. The queried neuro-SNs respond to the reader with impulse radio (IR) transmissions that directly encode the sensed local events. The reader processes these event-driven signals to determine whether the monitored environment is in a normal or anomalous state, while rigorously controlling the false discovery rate (FDR) of detections below a predefined threshold. The proposed approach employs an online hypothesis testing method with e-values to maintain FDR control without requiring knowledge of the anomaly rate, and it dynamically optimizes the sensor querying strategy by casting it as a best-arm identification problem in a multi-armed bandit framework. Extensive performance evaluation demonstrates that the proposed method can reliably detect anomalies under stringent FDR requirements, while efficiently scheduling sensor communications and achieving low detection latency.
\end{abstract}

\section{Introduction}

Neuromorphic technologies – encompassing sensing \cite{lichtsteiner2006128, martini2022lossless}, computing \cite{davies2018loihi, liu2014event}, and communication \cite{chen2023neuromorphic, skatchkovsky2020end, chen2024neuromorphic, chen2022neuromorphic, 10946192, lee2024asynchronous} -- are emerging as a promising paradigm for ultra-low-power, low-latency sensor networks. Inspired by the brain's event-driven information processing, a neuromorphic sensor  records information by producing spikes at times marking significant changes in the monitored environment \cite{schaefer2022aegnn}. Thanks to the inherent capacity of neuromorphic sensors to respond only to relevant events, incorporating neuromorphic sensors in a remote monitoring system can drastically reduce redundant transmissions, leading to substantial energy and bandwidth savings and supporting the deployment of large populations of wireless sensors \cite{chen2023neuromorphic, lee2024asynchronous}. 

\begin{figure}[t]
\centering
\includegraphics[width=0.45\textwidth]{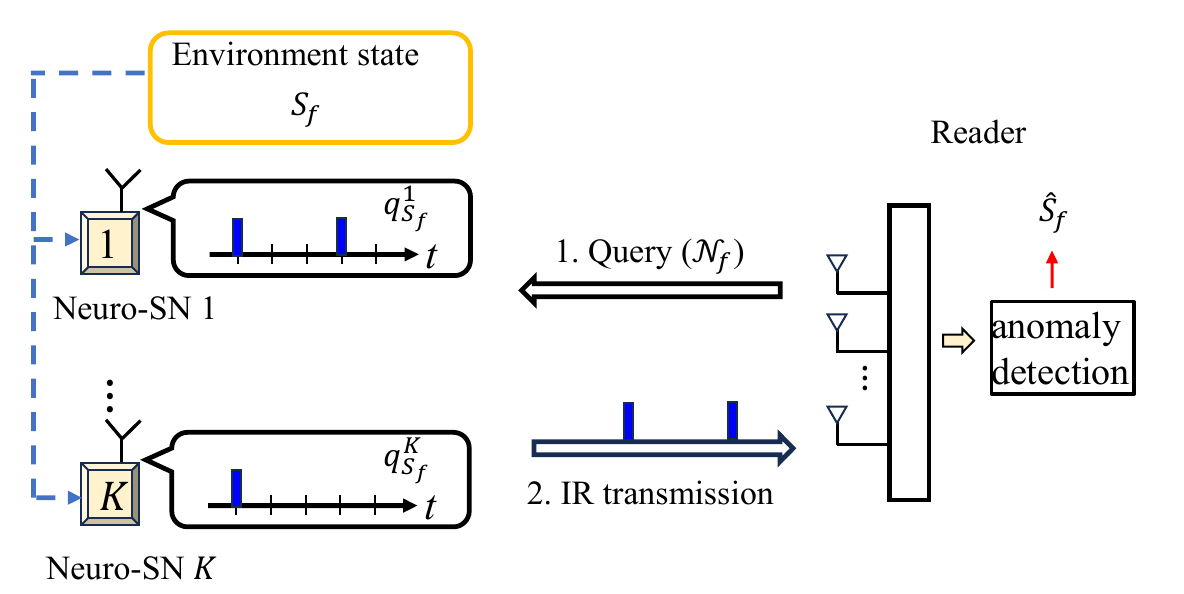}
\caption{In the setup under study, a reader interrogates a subset $\mathcal{N}_f$ of the $K$ neuromorphic sensor nodes (neuro-SN) at each frame $f$ in order to determine whether the behavior of the monitored system is normal or anomalous. The binary variable $S_f$ indicates the presence or absence of an anomaly at frame $f$, while $\hat{S}_f$ represents the corresponding estimate of the reader.  }
\label{Fig:basic setup}
\end{figure}

For example, reference \cite{lee2024asynchronous} demonstrated an asynchronous wireless network of salt-grain-sized sensors that mimic neural spikes, showing that hundreds of microscale nodes can communicate sporadic bursts of data to a central receiver using backscattering. Such neuromorphic networks have been highlighted for their potential in brain–machine interfaces, where implantable sensors would relay neural spikes in real time, as well as in other domains like wearable devices and environmental sensing \cite{spectrum}.

In applications such as remote surveillance and autonomous systems, timely detection of abnormal events is critical. Notably, neuromorphic vision sensors can detect sudden motion or the appearance of an intruder -- e.g., an unauthorized drone -- by producing rapid spikes, while remaining largely inactive under normal conditions \cite{lichtsteiner2006128}. In environmental monitoring, distributed sensors might pick up abrupt changes in temperature or smoke indicative of a forest fire \cite{alkhatib2014review}. In these scenarios, the system must provide reliable anomaly detection despite limited energy supply and communication bandwidth. 

To this end,  the detection algorithm needs to be statistically robust,  minimizing false alarms to avoid wasting attention and energy on benign events. Furthermore, the network must intelligently decide which sensors to query and when, so as to maximize the chance of detecting an anomaly while conserving resources. In this regard, a common reliability criterion is maintaining the \emph{\gls{fdr}} below a target level, ensuring that only a small fraction of reported anomalies are false positives \cite{rebjock2021online, ahmad2017unsupervised, lavin2015evaluating}. Achieving a low \gls{fdr} is especially important in safety-critical applications, like health monitoring or security,  to maintain trust in the system's alarms.

This paper addresses these challenges by proposing a novel solution that integrates tools from statistical FDR control \cite{javanmard2018online} and adaptive learning \cite{garivier2016optimal} into a neuromorphic wireless sensing framework.

\subsection{Related Work}

\emph{Neuromorphic sensing}: Neuromorphic sensing aims to emulate biological sensory systems by encoding information through sparse, event-driven signals rather than continuous streams. A prominent category is vision sensing, where \glspl{dvs} detect asynchronous changes in brightness at each pixel, achieving low latency and high energy efficiency. Prophesee and IniVation are among the leading companies in this area, offering state-of-the-art event-based cameras for machine vision and robotic applications \cite{event, event2}. Beyond vision, companies like SynSense are developing ultra-low-power neuromorphic processors such as Speck, which enable local event-driven processing for embedded intelligence \cite{event3}. 

\emph{Neuromorphic  communication}: Recent work~\cite{skatchkovsky2020end} has introduced end-to-end neuromorphic communication systems that replaces conventional sensing and frame-based radio with a fully spike-driven pipeline. In it, neuromorphic sensors produce asynchronous event spikes which are encoded by a \gls{snn} and transmitted as \gls{ir} pulses, then decoded by another \gls{snn} at the receiver. This all-spike design was shown to significantly lower latency and improve efficiency compared to conventional synchronous digital communication. 

Building on this, reference \cite{chen2023neuromorphic} extended the approach to a multi-user, fading-channel scenario. In it, each \gls{iot} device uses a neuromorphic event sensor, on-device \gls{snn}, and \gls{ir} transmitter, communicating over a shared wireless channel to a receiver that also employs an \gls{snn}. A key principle in these systems is event-driven data acquisition, which allows the wireless network to inherently focus on the “semantics” of the data – i.e. changes that carry new information – rather than sending redundant raw readings.

Further energy savings can be obtained by combining \gls{ir}-based communication with wake-up radio triggers~\cite{chen2024neuromorphic}, adopting different design criteria~\cite{ke2024neuromorphic}, and leveraging multi-level spikes~\cite{10946192}. In particular, reference~\cite{10946192} shows that associating each spike with a small payload can improving inference accuracy without increasing spike count. This study also includes an experimental evaluation using software-defined radios.

\emph{Online anomaly detection}: Online anomaly detection focuses on identifying abnormal patterns in streaming data in real time, without requiring access to the entire dataset beforehand. For example, reference~\cite{rebjock2021online} proposed online false discovery rate control methods for anomaly detection in time series to address the challenges of rare anomalies and temporal dependencies. Reference~\cite{ahmad2017unsupervised} proposed an unsupervised real-time anomaly detection algorithm for streaming data, which is capable of continuous learning and early detection of both spatial and temporal anomalies without requiring labeled data or batch processing.

\emph{Multi-armed bandit optimization}: \Gls{mab} optimization has been extensively studied in previous works. For example, reference 
\cite{garivier2016optimal} proposed the track-and-stop strategy for best-arm identification under fixed-confidence settings. Reference \cite{kaufmann2016complexity} studied the complexity of best-arm identification in stochastic MAB under both fixed-budget and fixed-confidence settings. To speed up the convergence of MAB in unseen tasks, \cite{zhang2023bayesian} proposed for the first time the use of contextual meta-learning that incorporates auxiliary optimization tasks knowledge.

\subsection{Main Contributions}

Building on the above context and prior art, this paper develops a protocol for online anomaly detection via neuromorphic sensor networks. The main contributions are summarized as follows:
\begin{itemize}
\item	\emph{Online anomaly detection framework based on neuromorphic sensors}: We introduce a novel online anomaly detection system based on neuromorphic sensor networks. As shown in Fig.~\ref{Fig:basic setup}, this model implements a pull-based communication strategy, where a central reader intermittently queries neuromorphic sensor nodes and receives event-driven \gls{ir} spike signals. The model explicitly incorporates the sparse, asynchronous nature of neuromorphic data and the requirement to control the false discovery rate in detection decisions.
\item	\emph{Adaptive \gls{fdr} control with e-values}: We design an adaptive statistical testing framework that guarantees FDR control at any desired level in an online setting. In contrast to traditional methods that rely on fixed p-value thresholds, our framework utilizes e-values \cite{ramdas2024hypothesis} to accumulate evidence over each frame. This allows the detector to flexibly adjust its confidence as new spike data arrive, ensuring that the expected proportion of false alarms remains below the target level even when the anomaly occurrence rate is unknown and nonstationary.
\item	\emph{Best-arm identification for sensor scheduling}: We introduce a sensor scheduling policy based on best-arm identification principles from multi-armed bandit theory \cite{garivier2016optimal}. At each time frame, the reader must decide which sensors to query next. Our policy dynamically learns the sensors that are most likely to yield an anomaly indication by treating each sensor as an ``arm' with an unknown reward. By continuously updating estimates and concentrating queries on the most promising sensor, the policy achieves efficient allocation of limited query opportunities, accelerating the detection of anomalies under uncertainty.
\item	 \emph{Extensive performance evaluation}: We provide a thorough performance evaluation of the proposed system through analysis and simulations. The results demonstrate that our approach can detect anomalies quickly and reliably under a variety of conditions. In particular, we show that the FDR is maintained below the target level in all tested scenarios, validating the effectiveness of the e-value based control. Moreover, the adaptive scheduling policy is shown to significantly reduce detection delay and resource usage compared to non-adaptive or random scheduling baselines. 
\end{itemize}

The rest of the paper is organized as follows. Section \ref{sec:sys} describes the pull-based neuromorphic wireless communications system, while the online anomaly detection based on hypothesis test framework is described in Section \ref{sec:online}. Section \ref{sec:node_selection} explains the scheduling policy optimization based on best-arm identification. Experimental setting and results are described in Section \ref{sec:exp}. Finally, Section \ref{sec:con} concludes the paper.

\section{System model}\label{sec:sys}
\subsection{Setting}
Fig.~\ref{Fig:basic setup} shows the basic setup considered in this paper. We consider a pull-based neuromorphic wireless communications scenario where the reader conducts a continuous online anomaly detection at each time frame based on the deployment of \gls{neurosn}~\cite{lee2024asynchronous}. 

The $K$ \glspl{neurosn} observe a common environment via local neuromorphic sensors \cite{chen2023neuromorphic}, and communicate with the reader via \gls{ir}. We define the set of node indices by $\mathcal{K} = \{1, 2, \ldots, K\}$. Through communication with the \glspl{neurosn}, the reader continuously monitors the environment to detect anomalies, while ensuring that the \gls{fdr} is maintained below a pre-determined level, denoted as $\alpha$. 
The \gls{fdr} measures the fraction of detected anomalies that do not correspond to actual anomalies. 
Specific applications include real-time forest fire detection~\cite{alkhatib2014review} or drone activity monitoring~\cite{chen2023neuromorphic} based on wireless sensor networks. 

As in \cite{rebjock2021online}, anomalies are defined as rare events that significantly deviate from expected patterns. In our setting, an anomaly causes neuromorphic sensors to produce spikes, also known as events, at a significantly higher rate than under normal conditions. For example, neuromorphic cameras produce spikes at times in which activity is recorded in the monitored scene~\cite{chen2023neuromorphic}. Thus, in the presence of significant activity, e.g., from unauthorized drones, the spike rate is increased.

\begin{figure}[t]
\centering
\includegraphics[width=0.45\textwidth]{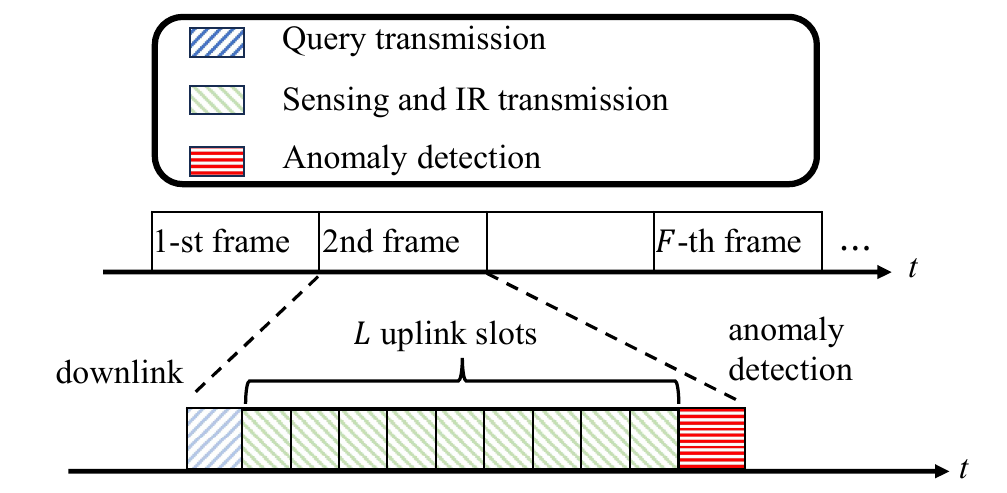}
\caption{Frame structure illustrating the query transmission from the reader (downlink), sensing and impulse radio (IR) transmission by the selected neuro-SN during $L$ uplink slots, and anomaly detection at the reader.}
\label{Fig:time}
\end{figure}

As illustrated in Fig.~\ref{Fig:time}, we adopt a discrete time model, in which time is divided into frames indexed by $f =1, 2, \ldots, F$. Each frame consists of three phases: 
\begin{enumerate}
\item{\textit{Downlink query signal transmission}: At the beginning of each frame, the reader first sends a query signal to a selected subset of devices.}
\item{\textit{Sensing and uplink \gls{ir} transmission}: Upon detecting the query signal, e.g., via a wake-up radio \cite{chen2024neuromorphic, shiraishi2024coexistence}, a \gls{neurosn} activates its neuromorphic sensor and its IR transmitter. The \gls{ir} transmission phase is divided into $L$ time slots, during which the \glspl{neurosn} sense and transmit in an online and event-driven manner.}
\item{\textit{Anomaly detection at the reader}: Finally, the reader determines whether the environment is in an anomalous state based on the received signals. }
\end{enumerate}
\subsection{Observation Models}\label{sec:observation_model}
Following~\cite{rebjock2021online}, we model the presence or absence of an anomaly in a frame indexed by $f$ as an arbitrary, unknown sequence of binary state $S_f$, which are defined as 
\begin{equation}
 S_f = \begin{cases} 0, & \text{no anomaly},\\1, & \text{anomaly}.\end{cases}
\end{equation}

\begin{figure}[t]
\centering
\includegraphics[width=0.45\textwidth]{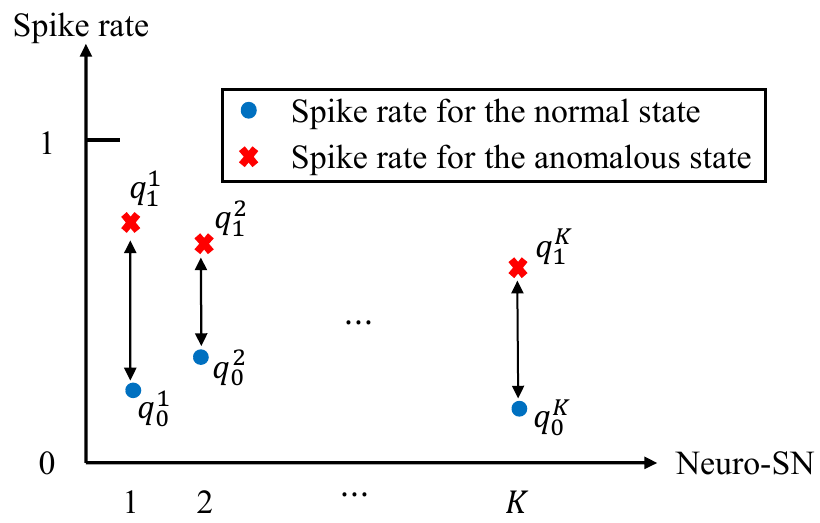}
\caption{Each $k$-th sensor is characterized by a known spiking probability $q^{k}_{0}$ in normal conditions, and a higher, unknown spiking probability $q^{k}_{1}\geq q^{k}_{0}$ in the presence of an anomaly.}
\label{Fig:source_model}
\end{figure}

We model the observations of each neuromorphic sensor $k$ at the $f$-th frame via a Bernoulli process with spiking rate determined by the current state $S_f$. Accordingly, for each sensor $k$, we set two spiking rates corresponding to the absence or presence of an anomaly, namely, $q^{k}_{0}$ and $q^{k}_{1}$. As illustrated in Fig.~\ref{Fig:source_model}, we assume the inequality $q^{k}_{0}~\leq~q^{k}_{1}$ for all $k~\in~\{1, 2, \ldots, K\}$.  This indicates that anomalies are characterized by larger spiking rates.

Accordingly, each node $k$ obtains a spike sequence $\bm{s}_{f}^{k} =[s_{f, 1}^{k}, \ldots, s_{f, L}^k] \in~\{0, 1\}^{L}$, where $s_{f, l}^k = 1$ indicates that node $k$ records a spike at time instant $l$ of the $f$-th frame, and $s_{f, l}^k = 0$ indicates that no spike was recorded. This means that we have
\begin{equation}
 s_{f, l}^k~\underset{\text{i.i.d.}}{\sim}~\mathrm{Bern}(q_f^k)~
 \text{with}~ q_f^k = 
 \begin{cases}q^{k}_{0}&\text{if}~S_f =0
 \\q^{k}_{1}&\text{if}~S_f =1.
 \end{cases}\label{eq:spike_generation}
\end{equation}

Through an initial calibration phase, the reader can obtain the statistics on the spike rates in normal conditions. Therefore, the reader is assumed to know the value of the probability $q^{k}_{0}$ for all $k~\in~\{1, 2, \ldots, K\}$. In contrast, we assume the reader does not know the spiking probability $q^{k}_{1}$ in the presence of anomalies.

\subsection{Scheduling and Impulse Radio Transmission}\label{sec:transmission_model}
In this work, we assume \gls{tdma}-based communication between the reader and \glspl{neurosn}, which consists of downlink phase and uplink phases. 
At each time frame $f$, the reader requests data from a subset $\mathcal{N}_f\subseteq \{1,\ldots,K\}$ of \gls{neurosn} by multicasting a query signal that encodes the target node IDs and its schedule information. 
The reader schedules $|\mathcal{N}_f|= C_\mathrm{max}$ nodes, and transmits this control information during the downlink phase. 
In practice, the cardinality $C_{\mathrm{max}}$ of the scheduled set depends on the available uplink capacity.

The nodes that receive the query signal begin sensing over the $L$ time instants within the frame. The spike sequence $\bm{s}_{f}^{k}$ is transmitted by all scheduled sensors in $\mathcal{N}_f$ in an online and event-driven manner using \gls{ir} by directly mapping a spike from the sensor to an electromagnetic pulse~\cite{chen2023neuromorphic}. 

\begin{figure}[t]
\centering
\includegraphics[width=0.45\textwidth]{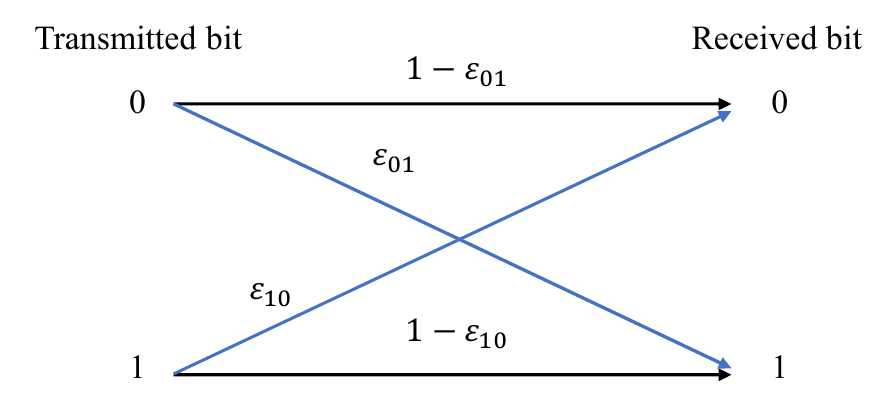}
\caption{Illustration of a binary asymmetric channel.}
\label{Fig:BASC}
\end{figure}

As illustrated in Fig.~\ref{Fig:BASC}, we model the channel as a binary asymmetric channel, where transmitted bits may be flipped at the receiver, with bit 0 flipping to 1 with probability $\epsilon_{01}$, and bit 1 flipping to 0 with probability $\epsilon_{10}$. The \gls{ir} receiver at the reader estimates the spike sequence transmitted by each \gls{neurosn}. After receiving the estimated spike sequence $\hat{\bm{s}}_{f}^{k} =[\hat{s}_{f, 1}^{k}, \ldots, \hat{s}_{f, L}^k]\in\{0,1\}^L$ for each node $k\in\mathcal{N}_f$ at time frame $f$, the reader uses this information to determine whether the current state $S_f$ is normal or anomalous, as described in the next section.

\section{Online Anomaly Detection} \label{sec:online}

In this section, we first present the hypothesis test framework, assuming that the reader has collected data from the subset of nodes $\mathcal{N}_f$ at the $f$-th frame. Then, we describe the proposed anomaly detection mechanism for online anomaly detection. The problem of optimizing node scheduling will be addressed in the next section.

\subsection{Hypothesis Testing Framework}\label{sec:hypothesis_testing}

At the end of each frame $f$, the reader tests whether the current system state is normal or anomalous. Specifically, it considers the following hypothesis
\begin{equation}
 \mathcal{H}_{f}: S_f = 0~\text{(normal state)}\label{hypo}
\end{equation} 
against the alternative hypothesis that there is an anomaly, i.e., $S_f = 1$. Equivalently, the reader evaluates the null hypothesis for the requested sensors in $\mathcal{N}_f$ as follows
\begin{equation}
 \mathcal{H}_{f}: \boldsymbol{s}_{f}^k \underset{\text{i.i.d.}}{\sim} \mathrm{Bern}(q^{k}_{0}),~\forall k\in\mathcal{N}_f
\end{equation}
against the composite alternative hypothesis
\begin{equation}
\bar{\mathcal{H}}_f: \boldsymbol{s}_{f}^k  \underset{\text{i.i.d.}}{\sim} \mathrm{Bern}(q^{k}_{1}),\text{for some}~ q^{k}_{1}>q^{k}_{0}, ~\forall k\in\mathcal{N}_f.
\end{equation}
In order to conduct a hypothesis test during the $f$-th frame, the reader uses the spike sequences $\hat{\bm s}_f=\{\hat{\bm s}_f^k\}_{k\in\mathcal{N}_f}$ observed during this frame and creates the test statistics $X_f$. In addition, the reader selects the rejection threshold $\alpha_f$ by considering the target metrics, such as \gls{fdr}. The null hypothesis $\mathcal{H}_f$ is that there is no anomaly, such that if this hypothesis is rejected, the reader declares an anomalous state. The decision at frame $f$ is denoted as $\hat{S}_f\in\{0,1\}$, which can be described as $\hat{S}_f = h(X_f,\alpha_f)$. Here, $h(X_f,\alpha_f)$ is an indicator function representing the rejection decision based on $X_f$ and $\alpha_f$. 
In the following subsection, we first describe our target metrics and use test statistics based on e-values. Finally, we describe how to select the rejection thresholds $\alpha_f$ considering the target metrics.

\subsection{Decaying Memory Time-Averaged False Discovery Rate}\label{sec: decayed_FDR}

Let $\mathcal{F}_{f| 0}$ represent the subset of the frame indices $\{1, 2, \ldots, f\}$ for which the environment is in a normal state, i.e., $\mathcal{F}_{f| 0} = \{f^{\prime} \in \{1, 2, \ldots, f\}: S_{f^{\prime}} = 0\}$. 
At frame $f$, the cumulative number of anomalies declared by the reader, averaged with a forgetting factor $0 < \delta < 1$ over the past frames, is 
\begin{align}
    \hat{A}_f = \sum_{f^{\prime}=1}^f \delta^{f-f^{\prime}}  \mathbbm{1}(\hat{S}_{f^{\prime}}=1), \label{R_f}
\end{align}
where $\mathbbm{1}(\cdot)$ is the indicator function, which equals to 1 if the argument is true, and equals to 0 otherwise. In a similar way, the weighted time-average of incorrectly detected anomalies is
\begin{align}
    F_{f} = \sum_{f^{\prime}=1}^f \delta^{f-f^{\prime}}  \mathbbm{1}(\hat{S}_{f^{\prime}}=1) \mathbbm{1}(S_{f^{\prime}}=0).
\end{align}

We are interested in controlling the decaying-memory time-averaged \gls{fdr}, which is defined as the ratio~\cite{rebjock2021online}
\begin{align}
\text{FDR}_f=\mathbb{E}\bigg[\frac{F_{f}}{\max(\hat{A}_f, 1)} \bigg], \label{fdrl}
\end{align}
This quantity measures the fraction of estimated anomalies that were actually normal states. Thus, controlling the \gls{fdr} in \eqref{fdrl} ensures that the system does not cause an excessive fraction of false alarms. Accordingly, the design target is to determine a sequence of rejection thresholds $\{\alpha_f\}_{f=1}^{\infty}$ ensuring that the metric $\text{FDR}_f$ is controlled below a pre-determined level $\alpha$, as 
\begin{equation}
\mathrm{FDR}_f \leq \alpha,~\text{for all frames}~f= 1, \ldots, F. \label{eq:FDR_condition}
\end{equation}

\subsection{Decaying Memory Time-Averaged True Discovery Rate}\label{sec:TDR}
The bound in \eqref{eq:FDR_condition} can be trivially guaranteed by setting $\alpha_f = 0$ for all frames $f$. In this case, no anomalies are detected, and thus there is no false detection of anomalies. However, this approach clearly fails to detect any potential anomalies. Therefore, a testing strategy must be also evaluated in terms of its ability to detect true anomalies. To account for this, we consider the \gls{tdr} as the average fraction of truly anomalous states that are correctly detected.

In a similar way to the decaying-memory time-averaged \gls{fdr}, let $A_f$ denote the cumulative number of anomalies that occur in the environment at frame $f$, averaged with a forgetting factor $0 < \delta < 1$, which is defined as
\begin{equation}
 A_f = \sum_{f^{\prime} = 1}^{f} \delta^{f- f^{\prime}}\mathbbm{1}(S_{f^{\prime}}=1).
\end{equation}
Similarly, the weighted time-averaged of correctly detected anomaly, denoted as $T_f$ can be expressed as
\begin{equation}
T_f =\sum_{f^{\prime}=1}^f  \delta^{f-f^{\prime}}\mathbbm{1}(\hat{S}_{f^{\prime}}=1) \mathbbm{1}(S_{f^{\prime}}=1).
\end{equation}
Finally, we define decayed-memory time-averaged \gls{tdr}, which is defined as the ratio, as described below:
\begin{equation}
    \mathrm{TDR}_f = \mathbb{E}\left[\frac{T_f}{\max(A_f, 1)}  \right].\label{eq:Def_TDR}
\end{equation}
This quantity measures how many actual anomaly events the reader successfully detects. This depends on the scheduling policy, which will be described in Sec.~\ref{sec:node_selection}.

The metrics \gls{fdr} in \eqref{fdrl} and \gls{tdr} in \eqref{eq:Def_TDR} will be used as a basis for designing rejection threshold $\alpha_f$. Specifically, in each frame, the reader sets a rejection threshold $\alpha_f$ based on the past rejection history so that  \gls{fdr} condition in \eqref{eq:FDR_condition} is always guaranteed while providing high \gls{tdr}. Specific algorithms should be designed by taking into account the well-known piggybacking and $\alpha$ death problem~\cite{rebjock2021online}, whose details are specified in Sec.~\ref{sec: design}. 
\subsection{Construction of E-Values}
An e-value $e_f \in [0, \infty]$ is a function of the current observations $S_f$ that satisfies the inequality $\mathbb{E}[e_f]~\leq~1$ when the null hypothesis $\mathcal{H}_f$ is true, i.e., when the state is normal ($S_f = 0$)~\cite{ramdas2024hypothesis}.  
To construct an e-value $e_f$ for each time frame $f$, the reader estimates the spike rate from node $k$ at the $f$-th frame as
\begin{align}
    \hat{q}^{k}_{f}=\frac{1}{L}\sum_{l=1}^L \hat{s}_{f,l}^k.\label{eq:estimaged_spike_rate_f}
\end{align}
Intuitively, a larger estimated rate $\hat{q}^k_f$ suggests a higher likelihood that the environment is experiencing an anomaly.

Since the anomalous spike rates are unknown to the reader, an e-value can be constructed by relying on the plug-in method~\cite{ramdas2024hypothesis} along with the merging of e-values for the scheduled sensors $k \in \mathcal{N}_f$. Specifically, for each sensor $k \in \mathcal{N}_f$, an e-value is obtained via the likelihood ratio
\begin{equation}
e_f^k = \left(\frac{\hat{\psi}^{k}_{f|1}}{\psi^{k}_{f|0}}\right)^{L\hat{q}^{k}_{f}}\left(\frac{1-\hat{\psi}^{k}_{f|1}}{1-\psi^{k}_{f|0}}\right)^{L(1-\hat{q}^{k}_{f})},\label{eq:e_value_channel}
\end{equation}
where
\begin{equation}
\psi^{k}_{f|0} = q^{k}_{0}(1-\epsilon_{10}) +(1- q^{k}_{0})\epsilon_{01}
\end{equation}
is the spiking probability that the reader receives from the \gls{neurosn} $k$ at the $f$-th frame when the environment is normal state and $\hat{\psi}^{k}_{f|1}$ is the maximum likelihood of received spike rate at the reader from \gls{neurosn} $k$ at the $f$-th frame when the environment is anomalous state. 
This is obtained as  
\begin{equation}
\hat{\psi}^{k}_{f|1} = \hat{q}^{k}_{1}(1-\epsilon_{10}) +(1- \hat{q}^{k}_{1})\epsilon_{01},
\end{equation}
where $\hat{q}^{k}_{1}$ is the maximum likelihood estimate of the spike rate $q^{k}_{1}$, which is given by
\begin{equation}
\hat{q}^{k}_{1} = \max(q^{k}_{0},~\hat{q}^{k}_{f}).
\end{equation}

Finally, the reader merges the e-values $\{e_f^k\}_{k \in \mathcal{N}_f}$ to obtain a single e-value via the arithmetic mean~\cite{ramdas2024hypothesis}

\begin{equation}
 \bar{e}_f(\mathcal{N}_f)=  \frac{1}{|\mathcal{N}_{f}|}\sum_{k \in \mathcal{N}_{f}} e_f^k.\label{eq:e_value}
\end{equation}

\subsection{Design of the Rejection Thresholds} \label{sec: design}
In order to ensure \gls{fdr} condition~\eqref{eq:FDR_condition}, we adopt the algorithm detailed in~\cite{rebjock2021online}. Accordingly, at the end of each time frame $f$, the reader dynamically controls the decision threshold $\alpha_{f}$ based on the past rejection history. 
An anomaly is declared if the statistic $e_f$ is larger than a threshold $1/\alpha_f$, i.e.,
\begin{align}
    \hat{S}_f= \mathbbm{1}\left(e_f > \frac{1}{\alpha_f}\right).\label{eq:test_estimation}
\end{align}
To describe the update of the threshold $\alpha_f$, let $\rho_{j}$ denote the time of $j$-th detected anomaly, i.e., 
\begin{equation}
    \rho_{j}=\min\left\{f ~\geq ~ 0 \Bigg| \sum_{f^{\prime} = 1}^f \mathbbm{1}\left(e_{f^{\prime}} > \frac{1}{\alpha_{f^{\prime}}}\right) \geq j\right\}, \label{rho}
\end{equation}
and let $\{\gamma_{f}\}_{f=1}^{\infty}$ denote a non-increasing sequence summing to $1$, where $\gamma_f=0$ for all $f\leq 0$.
We specifically use the default sequence~\cite{rebjock2021online}
\begin{equation}
    \gamma_{f} \propto \frac{\log{\left(\max(f, 2)\right)}}{f \exp\left({\sqrt{\log(f)}}\right)}. \label{gamma}
\end{equation}
Finally, let $\tilde{\gamma}_f = \max(\gamma_f, 1-\delta)$ as in~\cite{rebjock2021online}.
Then, the rejection threshold $\alpha_f$ is updated as
\begin{equation}
\alpha_{f} = {\alpha} \,\eta \,\tilde{\gamma}_f+ \alpha\sum_{j \geq 1}\delta^{f- \rho_{j}}\gamma_{f-\rho_{j}},\label{eq:alpha_update_rule}
\end{equation}
where $\eta$ is smoothing parameter $\eta > 0$. 

\begin{proposition}
[\cite{rebjock2021online}] {When the decision threshold $\alpha_f$ is updated according to \eqref{eq:alpha_update_rule} at each frame $f$, the anomaly detection rule in \eqref{eq:test_estimation} ensures that the decaying-memory time-averaged FDR defined in \eqref{fdrl} is controlled at level $\alpha$ for all frames $f$, thus satisfying condition \eqref{eq:FDR_condition}. }\label{proposition:FDR}
\end{proposition}

\section{Scheduling Policy Optimization}\label{sec:node_selection}
In order to increase the \gls{tdr} while maintaining \gls{fdr} control as in \eqref{eq:FDR_condition}, an ideal scheduling policy would select at each frame $f$ the subset $\mathcal{N}_f$ of sensors $k$ with the highest spiking rates $q^{k}_{1}$ under an anomalous state. In fact, this would maximize the chance of correctly detecting an anomaly via the test \eqref{eq:test_estimation}. However, the true spiking rates $q^{k}_{1}$, for $k \in \mathcal{K}$, are not known by the reader. Furthermore, anomalies occur arbitrarily and unpredictably, making it challenging to estimate the probabilities $q^{k}_{1}$. 

In this section, we address the problem of optimizing the set $\mathcal{N}_f$ by tackling these challenges. To start, we study the problem of identifying the best sensor 
\begin{align}
    k^*= \argmax_{k\in \mathcal{K}}~q^{k}_{1},\label{eq:BAI_Problem}
\end{align}
thus concentrating on the case $C_{\mathrm{max}} = |\mathcal{N}_f| =1$, in which the uplink channel supports the transmission of a single \gls{neurosn}. In this context, formally, the goal in this section is to minimize the number of frames used to identify the best \gls{neurosn} $k^*$ with a probability of error not exceeding a threshold $\delta_s \in (0, 1)$. 
An extension to any uplink capacity $C_{\max} \geq 1$ will be also provided at the end of this section. 

\subsection{Multi-Armed Bandit Formulation}\label{sec:MAB}
To facilitate the design of a scheduling policy, we make the working assumption that the anomaly process is i.i.d. across the frame index $f$, with an unknown anomalous probability $\pi_1$, i.e., $S_f \underset{}{\sim}\mathrm{Bern}(\pi_1)$. Furthermore, we also assume that the normal spiking probabilities of all \glspl{neurosn} are equal, i.e., 
\begin{equation}
 q^{k}_{0} = q_0,\label{eq:q_0_assumption}
\end{equation}
for all $k =1,\ldots, K$. Under these simplified assumptions, the spike rates $\frac{1}{L}\sum_{l=1}^{L}s_{f, l}^k$ of each \gls{neurosn} $k$ is i.i.d. across frames $f$, with an average

\begin{equation}
\mu^k = (1-\pi_1)q_0 + \pi_1q^{k}_{1}. \label{eq:mean_k}
\end{equation}

This implies that the problem \eqref{eq:BAI_Problem} of identifying the best sensor becomes equivalent to the problem of finding a node whose average spike rate $\mu^k$ is the highest. i.e., 
 \begin{align}
    k^*= \argmax_{k\in \mathcal{K}}~\mu^k.\label{eq:BAI_muk}
\end{align}
This optimization can be formulated as the fastest best arm identification problem, studied in~\cite{garivier2016optimal}. As demonstrated therein, a strategy known as track-and-stop reviewed next, is optimal.

\subsection{Track-and-Stop Strategy}
At each  $f$-th frame, the reader estimates the spike rate $\mu^k$ of \gls{neurosn} $k$ as
\begin{equation}
  \hat{\mu}^{k}_{f} = \frac{1}{N^{k}_{f}}\sum_{f^{\prime} =1}^f \hat{q}_{f^{\prime}}^{k} \mathbbm{1}\left(k \in \mathcal{N}_{f^{\prime}}\right),\label{eq:maximum_likelihood}
\end{equation}
with
\begin{equation}
N^{k}_{f}= \sum_{f^{\prime} = 1}^f \mathbbm{1}\left(k \in \mathcal{N}_{f^{\prime}}\right)
\end{equation}
being the number of frames in which \gls{neurosn} $k$ was scheduled and with $\hat{q}_{f^{\prime}}^{k}$ defined in \eqref{eq:estimaged_spike_rate_f}. 

Define as ${\bm{U}}_f = \{k: N^{k}_{f} < \sqrt{f} - \frac{K}{2}\}$ the subset of nodes that have been scheduled fewer than $\sqrt{f} - \frac{K}{2}$ times up to frame $f$. If this set is not empty, the track-and-stop strategy selects one of the \gls{neurosn}s that was scheduled least frequently, i.e., in the smallest number of frames, emphasizing exploration~\cite {garivier2016optimal}. Otherwise, the scheduling selection accounts also for the current spiking rate estimates $\hat{\bm{\mu}}_f = \{ \hat{\mu}^{1}_{f}, \hat{\mu}^{2}_{f}, \ldots, \hat{\mu}^{K}_{f}\}$, considering both exploitation and exploration. Overall, the scheduling decision is
\begin{equation}
n_f= \begin{cases} \underset{k \in \mathcal{K}}{\argmin}~N^{k}_{f},& \text{if}~|\mb{U}_f| > 0,\\
\underset{k \in \mathcal{K}}{\argmin}~ \big\{N^{k}_{f}-f \cdot w_{\sigma({k})}({\hat{\bm{\mu}}}_f)\big\}, & \text{otherwise},
\end{cases}\label{eq: D_sampling}
\end{equation}
where $\sigma: \{1, 2, \ldots, K\} \rightarrow \{1, 2, \ldots, K\}$ is a function that returns the rank of each node $k$ in terms of the estimated spike rate $\hat{\mu}^k$ in descending order, satisfying the inequalities $\hat{\mu}_f^{\sigma(1)} \geq \hat{\mu}_f^{\sigma(2)} \ldots \geq \hat{\mu}_f^{\sigma(K)}$. The function $w_{\sigma(k)}(\hat{\bm{\mu}}_f)$, which returns positive value, is detailed in~\cite{garivier2016optimal} and satisfies the following property is true: $w_{1}(\bm{\mu}) \geq \ldots \geq w_{K}(\bm{\mu})$. This way, \gls{neurosn}s with a larger estimated spiking rate are prioritized by the selection rule \eqref{eq: D_sampling}.

\subsection{Extension of the Scheduling for Any Scheduling Capacity}
In this subsection, we extend the track-and-stop strategy to the case where it is possible to schedule $C_{\mathrm{max}} = |\mathcal{N}_f| \geq 1$ \glspl{neurosn}.  
To this end, we propose that, at the beginning of each frame, the reader selects $C_{\mathrm{max}} = |\mathcal{N}_f|$ nodes for scheduling by recursively applying the track-and-stop method in~\eqref{eq: D_sampling}. Specifically, the reader selects the scheduled nodes one-by-one, obtaining the set $\mathcal{N}_f = \{n_f^i\}_{i= 1}^{C_{\mathrm{max}}}$. In more detail, define $\bm{U}_f^i = \{k \in \mathcal{K}\setminus \{n_f^j\}_{j=1}^{i-1}: N^{k}_{f} < \sqrt{f} - \frac{K}{2}\}$ as the set of \glspl{neurosn} not yet included in the set $\mathcal{N}_f$ that have been scheduled in the past frames fewer than $\sqrt{f} - \frac{K}{2}$ times. Then, the $i$-th scheduling node at the $f$-th frame is selected by applying the rule \eqref{eq: D_sampling} as
\begin{equation}
n_f^i= \begin{cases} \underset{k \in \mathcal{K} \setminus \{n_f^j\}_{j = 1}^{j = i-1}}{\argmin}~N^{k}_{f},& \text{if}~|\bm{U}_f^i| > 0,\\
\underset{k \in \mathcal{K} \setminus \{n_f^j\}_{j = 1}^{j = i-1}}{\argmin}~ N^{k}_{f}-f w_{\sigma(k)}({\bm{\hat{\mu}}}_f), & \text{otherwise},
\end{cases}\label{eq: D_sampling_Extend}
\end{equation}
where function $w_{\sigma(k)}(\cdot)$ is as defined in \eqref{eq: D_sampling}.

\section{Numerical Results}\label{sec:exp}
In this section, we present numerical results to validate the performance of the proposed online anomaly detection system.
\subsection{Simulation Setting}
 In our experiments, the anomaly process $S_f$ is i.i.d. across all frames $f$, so that at each frame $f$ the environment is in the anomalous state with probability $\pi_1$ and in the normal state with probability $1 -\pi_1$, independently from frame to frame. We simulate a setting in which, as in~\eqref{eq:q_0_assumption}, the spike rate $q^{k}_{0}$ of a normal state is the same across all nodes, i.e., $q^{k}_{0} = q_0$ for all $k \in \mathcal{K}$. In contrast, the spiking rate $q^{k}_{1}$ under the anomalous state is uniformly generated within the interval $[q_0, q_0 +\Delta_{\mathrm{max}}]$ for each \gls{neurosn} $k$. Table \ref{table:para} shows our default parameters setting, which are used throughout the experiments unless stated otherwise.

\begin{table}[htp]
\caption{Default parameter setting.}
\begin{center}
\begin{tabular}{|c|c| p{10zw}} \hline
Parameters & Values  \\ \hline 
Number of \gls{neurosn} $K$ & 5\\
Forgetting factor $\delta$ & 0.99~\cite{rebjock2021online} \\
$\eta$& 0.99 \\
Number of uplink slots $L$& 50 \\
Target FDR $\alpha$ & 0.1~\cite{rebjock2021online} \\
Simulation runs & $10^3$  \\ 
Spiking rate under the normal state $q_0$ & $0.10$  \\ 
Anomaly ratio $\pi_1$ & 0.05\\ 
Error rates ($\epsilon_{01}$, $\epsilon_{10}$)&(0, 0)\\  
Upper bound on the spiking probability $\Delta_{\mathrm{max}}$&0.5\\ 
Channel capacity $C_{\mathrm{max}}$& 1\\ \hline
\end{tabular}
\label{table:para}
\end{center}
\end{table}
\subsection{On the Validity of Online Anomaly Detection}

We first investigate the validity of the online anomaly detection algorithm introduced in Sec.~\ref{sec:online}. To this end, we compare the performance of the proposed scheme with adaptive threshold $\alpha_f$ described in~\eqref{eq:alpha_update_rule} with a more conventional fixed-threshold strategy, in which we apply the threshold $\alpha_f = \alpha$ for all $f \in [1, F]$. Note that this combined approach does not satisfy the theoretical statistical guarantee in~\eqref{eq:FDR_condition}, and its performance can only be evaluated via numerical results.

\begin{figure*}[t]
    \centering
    \input{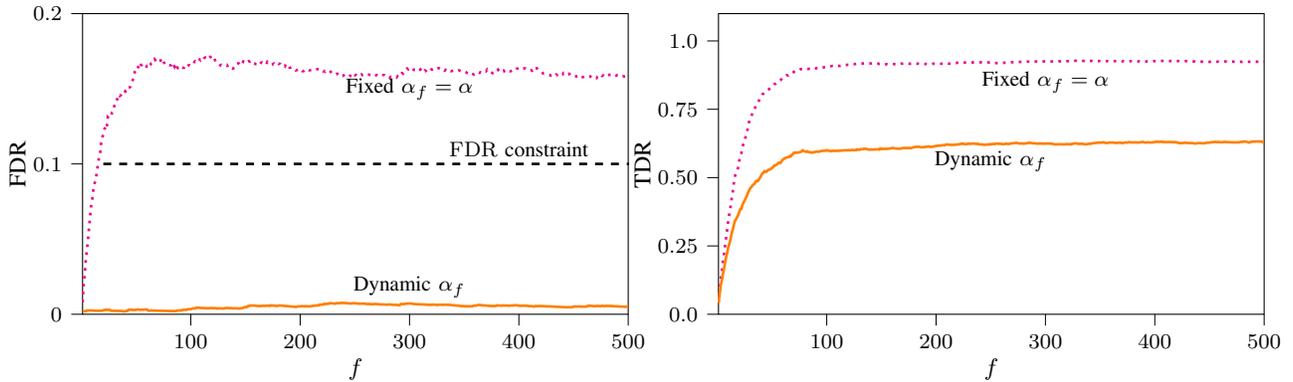}
     \caption{\gls{fdr} and \gls{tdr} versus the time frame $f$ for fixed anomaly decision threshold $\alpha_f = \alpha$ for all $f \in [1,F]$ and for the proposed dynamic anomaly decision threshold based on~\eqref{eq:alpha_update_rule}, where we set $K =1$ and $\pi_1 = 0.05$.}
    \label{fig:comparison_anomaly_detection}
\end{figure*}

Fig.~\ref{fig:comparison_anomaly_detection} shows \gls{fdr} and \gls{tdr} versus the time frame $f$ for fixed anomaly threshold method and for the proposed dynamic anomaly detection method in~\eqref{eq:alpha_update_rule}, where we set the number of \gls{neurosn} $K = 1$. As anticipated, the fixed-threshold is not able to always satisfies the \gls{fdr} constraint in~\eqref{eq:FDR_condition}. This is because this approach does not take into account the past rejection history for the current anomaly decision tasks. A fixed threshold can achieve a higher \gls{tdr} than the proposed method, but only by violating the \gls{fdr} constraint in~\eqref{eq:FDR_condition}. 
In contrast, the introduced methodology, dynamic threshold always satisfies the \gls{fdr} constraint $\alpha$, validating Proposition~\ref{proposition:FDR}.

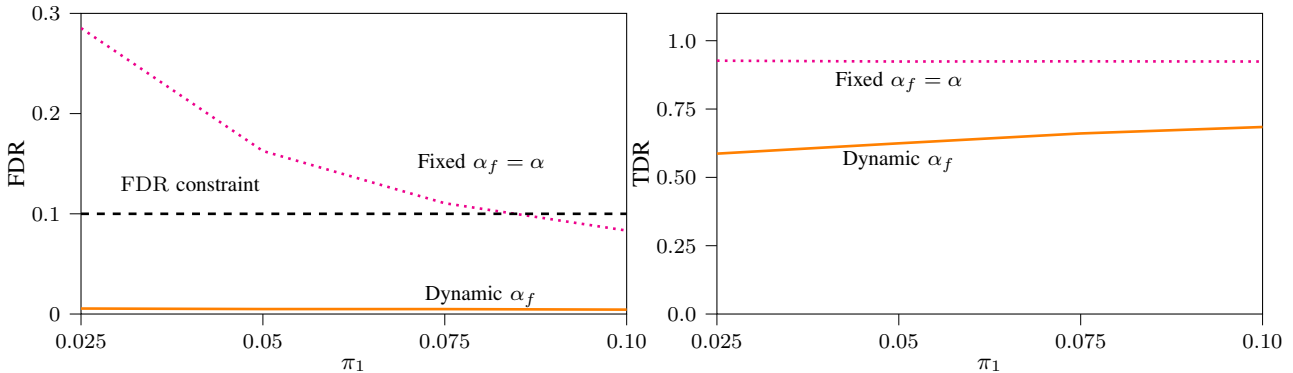
\begin{figure*}[t]
    \centering
    \begin{tikzpicture}

\definecolor{crimson2143940}{RGB}{214,39,40}
\definecolor{darkgray176}{RGB}{176,176,176}
\definecolor{darkorange25512714}{RGB}{255,127,14}
\definecolor{forestgreen4416044}{RGB}{44,160,44}
\definecolor{lightgray204}{RGB}{204,204,204}
\definecolor{mediumpurple148103189}{RGB}{148,103,189}

\def\hsep{1.2cm}
\def\vsep{1cm}
\def\vside{4cm}
\def\hside{0.4\textwidth}

\begin{groupplot}[
group style={group name=system, group size=3 by 1,  horizontal sep=\hsep, vertical sep=\vsep, xlabels at=edge bottom}, 
title style={at={(0.5,0.85)}},
anchor=south east,
height=\vside,
width=\hside,
scale only axis,
legend style={  
  at={(0, 1.05)}, 
  draw=none,
  fill opacity=0,
  anchor=south,  
  /tikz/every even column/.append style={column sep=0.3cm}
},
legend columns=2,
xmin=0.025, xmax=0.1,
xtick={0.025, 0.05, 0.075, 0.1},
xticklabels={
  \(\displaystyle {0.025}\),
  \(\displaystyle {0.05}\),
  \(\displaystyle {0.075}\),
  \(\displaystyle {0.10}\), 
},
xlabel={$\pi_1$},
xtick style={color=black},
scaled x ticks = false,
ymin=0.4, ymax=1,
ytick={0,0.25,0.5,0.75,1,1.5,2.},
yticklabels={
  \(\displaystyle {0.0}\),
  \(\displaystyle {0.25}\),
  \(\displaystyle {0.50}\),
  \(\displaystyle {0.75}\),
  \(\displaystyle {1.0}\),
  \(\displaystyle {1.5}\),
  \(\displaystyle {2.0}\),  
},
ylabel shift=-4pt,
xlabel shift=-3pt,
]

\nextgroupplot[
ylabel={FDR},
ymin=0,
ymax=0.3,
  ytick={0, 0.1, 0.2, 0.3},
 yticklabels={
 \(\displaystyle 0\),
 \(\displaystyle 0.1\),
 \(\displaystyle 0.2\),
  \(\displaystyle 0.3\),
  },
]

\addplot[dotted,  color=magenta, line width=1pt ]
table {%
0.025	0.2853288554204578
0.05	0.1626905395328454
0.07500000000000001	0.11054924188057383
0.1	0.08328587825332331

};
\node[black, font=\footnotesize] at (axis cs:0.08,0.15) {Fixed $\alpha_f = \alpha$};

\addplot[solid,  color=orange, line width=1pt ]
table {%
0.025	0.0054799158020811594
0.05	0.004952257269430728
0.07500000000000001	0.004906771170183791
0.1	0.0043876513398738715

};
\node[black, font=\footnotesize] at (axis cs:0.08,0.018) {Dynamic $\alpha_f$};

\addplot[dashed, color=black, line width=1pt, mark size=1pt]
table {%
0.025 0.1
0.10 0.1
};
\node[font=\footnotesize] at (axis cs:0.04,0.13) {$\mathrm{FDR}$ constraint};

\nextgroupplot[
ylabel={TDR},
ymin=0.0,
ymax=1.1,
]

\addplot[dotted,  color=magenta, line width=1pt ]
table {%
0.025	0.9269033303217433
0.05	0.9238416866034885
0.07500000000000001	0.9245968677169897
0.1	0.9238346516545384

};
\node[black, font=\footnotesize] at (axis cs:0.05,0.85) {Fixed $\alpha_f = \alpha$};

\addplot[solid,  color=orange, line width=1pt ]
table {%
0.025	0.5868068878497541
0.05	0.624560255067178
0.07500000000000001	0.6606861532424358
0.1	0.6842653096223036

};
\node[black, font=\footnotesize] at (axis cs:0.05,0.56) {Dynamic $\alpha_f$};

\end{groupplot}

\end{tikzpicture}
     \caption{\gls{fdr} and \gls{tdr} versus the anomaly probability $\pi_1$ for fixed anomaly decision threshold $\alpha_f = \alpha$ for all $f \in [1,F]$ and for the proposed dynamic anomaly decision threshold based on~\eqref{eq:alpha_update_rule}, where we set $K =1$ and $F = 1,000$.}
    \label{fig:acc_anomaly_detection_against_pi1}
\end{figure*}

Fig.~\ref{fig:acc_anomaly_detection_against_pi1} shows the \gls{fdr} and \gls{tdr} as a function of the anomaly probability $\pi_1$, where we set $K = 1$ and $F = 1000$. We observe that the fixed-threshold approach suffers from the higher \gls{fdr} when the anomaly probability $\pi_1$ is small, while the proposed dynamic anomaly detection threshold method can always satisfy the \gls{fdr} constraint. 
When the anomaly probability $\pi_1$ is small, with a dynamic threshold, the reader takes a conservative decision by setting a relatively smaller threshold $\alpha_f$, will a fixed-threshold method cannot adjust to the anomaly probability.

\subsection{Random v.s. Track-and-Stop Scheduling}
\begin{figure*}[t]
    \centering
    \input{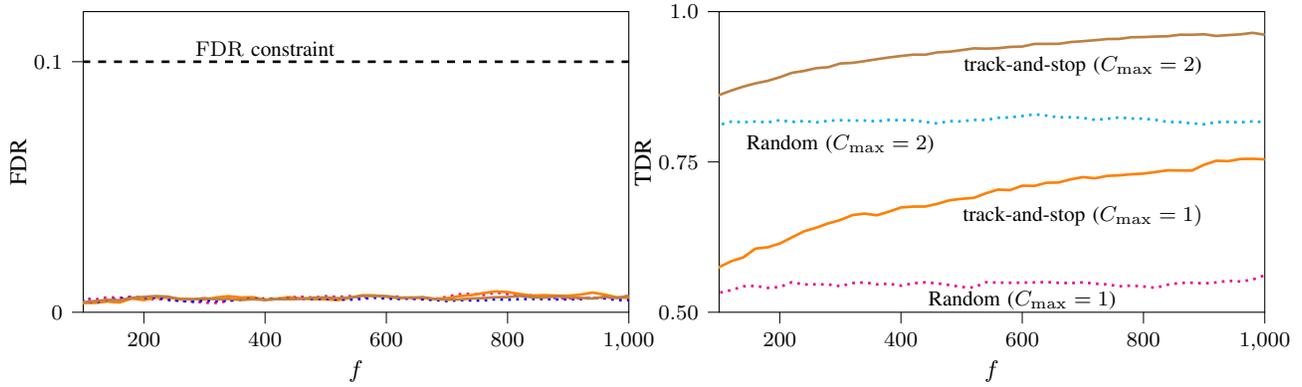}
     \caption{\gls{fdr} and \gls{tdr} versus the time frame $f$ for random sampling scheme and track-and-stop.}
    \label{fig:comparison}
\end{figure*}

We now study the impact of the scheduling strategy. To this end, Fig.~\ref{fig:comparison} shows the \gls{fdr} and the \gls{tdr} versus the time frame $f$ for the proposed anomaly detection scheme under a random scheduling strategy that selects $C_{\mathrm{max}}$ \glspl{neurosn} uniformly at random at every frame and for the track-and-stop method presented in Sec.~\ref{sec:node_selection}. We set the parameters described in Table~\ref{table:para} with the uplink channel capacity $C_{\mathrm{max}} = \{1, 2\}$ and anomaly probability $\pi_1 = 0.1$. 

All schedulers, when coupled with the proposed anomaly detection, are seen to satisfy the \gls{fdr} requirements, confirming the theoretical results in Proposition~\ref{proposition:FDR}. Moreover, the figure highlights the benefits of track-and-stop, increasing the \gls{tdr} for both uplink capacities $C_{\mathrm{max}} = 1$ and $C_{\mathrm{max}} = 2$. This advantage stems from the capacity of track-and-stop to select the \glspl{neurosn} that are most likely to contribute to the identification of the \glspl{neurosn} with highest spiking rates. 

From Fig.~\ref{fig:comparison} we also see that \gls{tdr} becomes larger as the number of scheduled nodes $C_{\mathrm{max}}$ increases, demonstrating the importance of collecting information from multiple \glspl{neurosn}. This is because by calculating e-value~\eqref{eq:e_value} based on the spike rate from multiple nodes, the reader can mitigate the negative impact of small e-value caused by the selection of nodes having small spike rate in anomalous state, $q_{1}^k$.

\subsection{Impact of System Parameters}
In this subsection, we focus on the performance of track-and-stop scheduling, evaluating the impacts of  different system parameters.
\subsubsection{On the \gls{fdr} constraint $\alpha$}
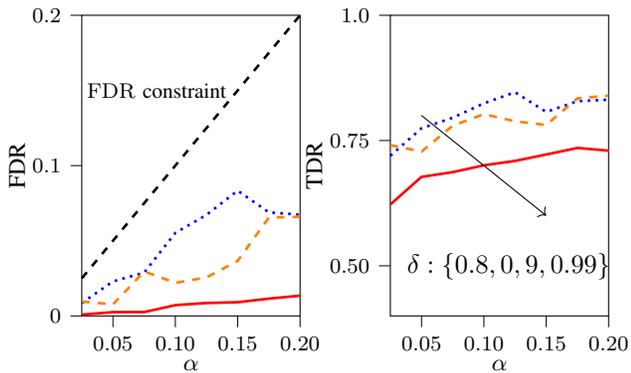
\begin{figure}[t]
    \centering
   \begin{tikzpicture}
\definecolor{crimson2143940}{RGB}{214,39,40}
\definecolor{darkgray176}{RGB}{176,176,176}
\definecolor{darkorange25512714}{RGB}{255,127,14}
\definecolor{forestgreen4416044}{RGB}{44,160,44}
\definecolor{lightgray204}{RGB}{204,204,204}
\definecolor{mediumpurple148103189}{RGB}{148,103,189}

\def\hsep{1.2cm}
\def\vsep{1cm}
\def\vside{4cm}
\def\hside{0.16\textwidth}

\begin{groupplot}[
group style={group name=system, group size=3 by 1,  horizontal sep=\hsep, vertical sep=\vsep, xlabels at=edge bottom}, 
title style={at={(0.5,0.85)}},
anchor=south east,
height=\vside,
width=\hside,
scale only axis,
legend style={  
  at={(1.2, 1.05)}, 
  draw=none,
  fill opacity=0,
  anchor=south,  
  /tikz/every even column/.append style={column sep=0.3cm}
},
legend columns=2,
xmin=0.025, xmax=0.20,
 xtick={0.05, 0.10, 0.15, 0.20},
xticklabels={
  \(\displaystyle {0.05}\),
  \(\displaystyle {0.10}\),
  \(\displaystyle {0.15}\),
  \(\displaystyle {0.20}\), 
},
xlabel={$\alpha$},
xtick style={color=black},
scaled x ticks = false,
ymin=0, ymax=1,
ytick={0,0.25,0.5,0.75,1,1.5,2.},
yticklabels={
  \(\displaystyle {0.0}\),
  \(\displaystyle {0.25}\),
  \(\displaystyle {0.50}\),
  \(\displaystyle {0.75}\),
  \(\displaystyle {1.0}\),
  \(\displaystyle {1.5}\),
  \(\displaystyle {2.0}\),  
},
ylabel shift=-4pt,
xlabel shift=-3pt,
]

\nextgroupplot[
ylabel={FDR},
ymin=0,
ymax=0.20,
  ytick={0, 0.1, 0.2},
 yticklabels={
 \(\displaystyle 0\),
  \(\displaystyle 0.1\),
  \(\displaystyle 0.2\),
  },
]
\addplot[dotted,  color=blue, line width=1pt ]
table {%
0.0250000000000000	0.00856357703578978
0.0500000000000000	0.0229793811528495
0.0750000000000000	0.0287749682148723
0.100000000000000	0.0552253230253158
0.125000000000000	0.0672315584549452
0.150000000000000	0.0833380773081787
0.175000000000000	0.0688711218779626
0.200000000000000	0.0673418684766597
};

\addplot[dashed,  color=orange, line width=1pt ]
table {%
0.0250000000000000	0.00960458382389353
0.0500000000000000	0.00766597463814724
0.0750000000000000	0.0295770391155169
0.100000000000000	0.0220158558243190
0.125000000000000	0.0256251416050864
0.150000000000000	0.0365961618210688
0.175000000000000	0.0654524124085955
0.200000000000000	0.0659218299329279
};
\addplot[solid,  color=red, line width=1pt ]
table {%
0.025	0.00084131788486741
0.05	0.002527429008694584
0.075	0.0025554223811669334
0.1	0.0071648962350101065
0.125	0.008587324180826635
0.15	0.009116927297175982
0.175	0.011494028147827231
0.2	0.013469469762409534
0.225	0.013938771460393658
0.25	0.017242094393239735

};



\addplot[dashed, color=black, line width=1pt, mark size=1pt]
table {%
0.025 0.025
0.25 0.25
};
\node[font=\footnotesize] at (axis cs:0.085,0.15) {$\mathrm{FDR}$ constraint};
\nextgroupplot[
ylabel={TDR},
ymin=0.4,
ymax=1,
]

\addplot[dotted,  color=blue, line width=1pt ]
table {%
0.0250000000000000	0.719574525471875
0.0500000000000000	0.774427167720550
0.0750000000000000	0.795513828650126
0.100000000000000	0.824265304934403
0.125000000000000	0.846427612048198
0.150000000000000	0.807108639176509
0.175000000000000	0.828499498574842
0.200000000000000	0.831547071419375
};

\addplot[dashed,  color=orange, line width=1pt ]
table {%
0.0250000000000000	0.740681527799719
0.0500000000000000	0.727888287746944
0.0750000000000000	0.779559240645187
0.100000000000000	0.802630394983807
0.125000000000000	0.788310046795058
0.150000000000000	0.781064776702811
0.175000000000000	0.834234214111214
0.200000000000000	0.839554511588868
};

\addplot[solid,  color=red, line width=1pt ]
table {%
0.025	0.6223175501417058
0.05	0.6774100344549487
0.075	0.6868983306099969
0.1	0.7005653090645748
0.125	0.7093292936102809
0.15	0.7221596239853563
0.175	0.7351776076637399
0.2	0.7298578642790483
0.225	0.7362807386498208
0.25	0.7281855061175501

};

\node at (axis cs:0.12,0.5) {$\delta: \{0.8, 0,9, 0.99\}$};
\draw[->] (axis cs:0.05,0.8) -- (axis cs:0.15,0.6);







\end{groupplot}




\end{tikzpicture}
     \caption{\gls{fdr} and \gls{tdr} versus the \gls{fdr} constraint $\alpha$ for track-and-stop (the forgetting factor $\delta = \{0.8, 0.9, 0.99\}$).}
    \label{fig:alpha_change}
\end{figure}

Fig.~\ref{fig:alpha_change} shows the \gls{fdr} and \gls{tdr} as a function of \gls{fdr} constraint $\alpha$. 
The figure highlights the trade-off between \gls{tdr} and \gls{fdr}: as the allocated \gls{fdr} constraint $\alpha$ becomes larger, the reader can more easily detect anomalies, increasing the \gls{tdr}.  We can also see that \gls{tdr} becomes larger as the forgetting factor $\delta$ becomes smaller. In fact, with a smaller $\delta$, the \gls{fdr} constraint puts higher weight on the results of the latest decisions, decreasing the importance of past false alarms. 

\subsubsection{On the sensitivity of the sensors}
\begin{figure}[t]
    \centering
    \begin{tikzpicture}
\definecolor{crimson2143940}{RGB}{214,39,40}
\definecolor{darkgray176}{RGB}{176,176,176}
\definecolor{darkorange25512714}{RGB}{255,127,14}
\definecolor{forestgreen4416044}{RGB}{44,160,44}
\definecolor{lightgray204}{RGB}{204,204,204}
\definecolor{mediumpurple148103189}{RGB}{148,103,189}

\def\hsep{1.2cm}
\def\vsep{1cm}
\def\vside{2.8cm}
\def\hside{0.18\textwidth}

\begin{axis}[%
width=7cm,
height=4cm,
scale only axis,
xmin=50,
xmax=500,
xlabel style={font=\color{white!15!black}},
xlabel={$f$},
ymin=0,
ymax=1,
scaled y ticks = false,
ylabel style={font=\color{white!15!black}},
ylabel={TDR},
axis background/.style={fill=white},
legend style={at={(0,1)},anchor=north west,legend cell align=left, align=left, draw=white!15!black},
legend columns=2
]

\addplot[dotted,  color=red, line width=1pt ]
table {%
20	0.0138887032372844
40	0.0155589553165495
60	0.0150547275931938
80	0.0151163610637453
100	0.0137126503348817
120	0.0118904688915290
140	0.0105899511311471
160	0.0114111440848446
180	0.0128558487299410
200	0.0131595573890805
220	0.0136121396451410
240	0.0138654169216397
260	0.0134247460736584
280	0.0129079873403453
300	0.0129919637412872
320	0.0127816302636461
340	0.0129843695719246
360	0.0121812492255919
380	0.0124381724787067
400	0.0124752575353257
420	0.0125585943281308
440	0.0131937383745580
460	0.0129674940394670
480	0.0126287899972002
500	0.0120269779834384
520	0.0119932335958041
540	0.0120351907546510
560	0.0125217753939849
580	0.0124844001089173
600	0.0115482082329011
620	0.0117377534330349
640	0.0124859409033960
660	0.0131540325250552
680	0.0129562220577641
700	0.0129961402863578
720	0.0135007266360915
740	0.0123460224751908
760	0.0112447711631582
780	0.0113321715989981
800	0.0121995327869010
820	0.0120872893734305
840	0.0120264063699805
860	0.0120954736815288
880	0.0113221108369267
900	0.0105433300973667
920	0.0112579519838746
940	0.0117974745698591
960	0.0117703552793046
980	0.0124052506421382
1000	0.0119885339797032
};
\node[black, font=\footnotesize] at (axis cs:400,0.05) {$\Delta_{\mathrm{max}} = 0.2$};

\addplot[dashed, color=blue, line width=1pt, mark size=1pt]
table {%
20	0.0920252530306125
40	0.121198518280041
60	0.123365922621143
80	0.129800084113967
100	0.131386169233003
120	0.134999590317316
140	0.136000570011318
160	0.140284638517535
180	0.145323840626088
200	0.140914667088201
220	0.139890388355452
240	0.142825149578682
260	0.142629390255510
280	0.142542446907042
300	0.145228123705959
320	0.145850563476093
340	0.145693043014356
360	0.149286509169613
380	0.150783250756035
400	0.155867153004752
420	0.157673258559368
440	0.159291743958417
460	0.163341076195481
480	0.162160539151031
500	0.163116593617409
520	0.162630898915006
540	0.166191381509529
560	0.165410359413992
580	0.166204799383275
600	0.170829160436731
620	0.171387731106511
640	0.169116380806509
660	0.170751536617790
680	0.172048308564372
700	0.174719329160389
720	0.178110518139874
740	0.180159824448364
760	0.178195524277289
780	0.180499599541611
800	0.180611289237269
820	0.184131763346037
840	0.185652163401176
860	0.185221699932727
880	0.180246064880519
900	0.182369376388864
920	0.182287409729024
940	0.180643799930013
960	0.176015663914045
980	0.178019620779378
1000	0.177101161978419
};
\node[black, font=\footnotesize] at (axis cs:300,0.2) {$\Delta_{\mathrm{max}} = 0.3$};

\addplot[ dashdotted, color=orange, line width=1pt, mark size=1pt]
table {%
20	0.215321094231145
40	0.310639340888134
60	0.342298507790211
80	0.358836674356744
100	0.379881459920768
120	0.379227576315965
140	0.395390319953146
160	0.402253155237332
180	0.405043041757529
200	0.417153019156675
220	0.420973950946213
240	0.434326636984845
260	0.438754170343236
280	0.441915526177613
300	0.453016957636939
320	0.459307940356216
340	0.466136494936152
360	0.465649124365444
380	0.465693596520521
400	0.465708741818166
420	0.470791767127099
440	0.471224771835118
460	0.467504466194063
480	0.470899802812409
500	0.473849055532278
520	0.473845411591916
540	0.471558353456249
560	0.477255546548046
580	0.477931893461125
600	0.481446208196266
620	0.486750599147678
640	0.484839193379580
660	0.489306939844732
680	0.493348593588087
700	0.492688986297841
720	0.492757905800774
740	0.496587439828219
760	0.495389382727569
780	0.502210555676099
800	0.507679896382466
820	0.512113179213335
840	0.509603335876185
860	0.513749603574422
880	0.513157571165741
900	0.515452698835920
920	0.517134962152650
940	0.513701544081932
960	0.515735640031673
980	0.523401022510191
1000	0.523842275563468
};
\node[black, font=\footnotesize] at (axis cs:350,0.4) {$\Delta_{\mathrm{max}} = 0.4$};

\addplot[ solid, color=cyan, line width=1pt, mark size=1pt]
table {%
20	0.344223429092729
40	0.477069648332176
60	0.528280241957225
80	0.555718959944127
100	0.571311744757448
120	0.580425514749149
140	0.588485335102679
160	0.598494216510547
180	0.612856323886187
200	0.619644395269338
220	0.626642962499938
240	0.642886188717064
260	0.649687718641035
280	0.652862701511101
300	0.659619214059597
320	0.667516273173848
340	0.671526567819769
360	0.671985187729842
380	0.679516278950811
400	0.685942086283428
420	0.694030612471945
440	0.696766621647512
460	0.699135161612460
480	0.704195547851507
500	0.707746457918566
520	0.704277730134862
540	0.709287078237183
560	0.709779100118515
580	0.714800986235623
600	0.718874185656701
620	0.718520120585326
640	0.725959343476756
660	0.727828353818611
680	0.732240666951062
700	0.738005220903238
720	0.736762700348808
740	0.741337195958181
760	0.734778189277990
780	0.738518221296821
800	0.743116843901195
820	0.741838394482745
840	0.747697810421713
860	0.756059001968133
880	0.758755616194841
900	0.762655220804072
920	0.760840468212284
940	0.756296808947111
960	0.763121805624109
980	0.769945848211712
1000	0.770428240207958
};
\node[black, font=\footnotesize] at (axis cs:350,0.75) {$\Delta_{\mathrm{max}} = 0.5$};

\end{axis}

\end{tikzpicture}
    \caption{\gls{tdr} versus the time frame $f$ for track-and-stop with different values of upper bound of sensor reading $\Delta_{\mathrm{max}}$.}
    \label{fig:sensitibitity_analysis_Delta_max}
\end{figure}
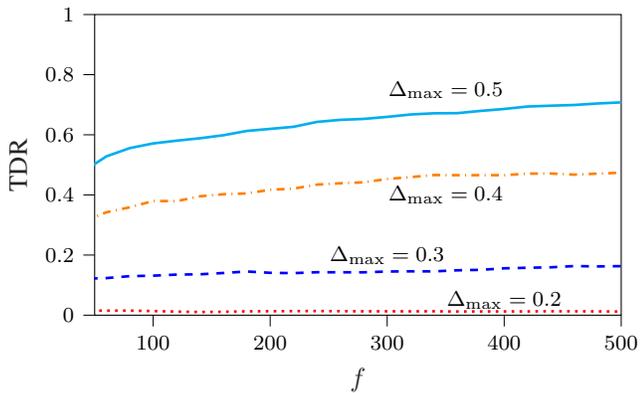

We now study the dependence of the system performance on the parameter defining the quality of the observations of sensors. First, we vary the upper bound $\Delta_{\mathrm{max}}$ on the spiking rate in the presence of anomalies. 
From Fig.~\ref{fig:sensitibitity_analysis_Delta_max}, we can see that \gls{tdr} is improved as the upper bound $\Delta_{\mathrm{max}}$ becomes larger, that is, the sensitivity improves. This is because a larger upper bound $\Delta_{\mathrm{max}}$ tends to increase the difference between the spike rate of \gls{neurosn} $k$ in the normal state $q^{k}_{0}$ and that in the anomalous state $q^{k}_{1}$, enabling the reader to detect anomalies with higher accuracy (see Fig.~\ref{Fig:source_model}).  

\begin{figure}[t]
    \centering
    \begin{tikzpicture}
\definecolor{crimson2143940}{RGB}{214,39,40}
\definecolor{darkgray176}{RGB}{176,176,176}
\definecolor{darkorange25512714}{RGB}{255,127,14}
\definecolor{forestgreen4416044}{RGB}{44,160,44}
\definecolor{lightgray204}{RGB}{204,204,204}
\definecolor{mediumpurple148103189}{RGB}{148,103,189}

\def\hsep{1.2cm}
\def\vsep{1cm}
\def\vside{2.8cm}
\def\hside{0.18\textwidth}

\begin{axis}[%
width=7cm,
height=4cm,
scale only axis,
xmin=50,
xmax=500,
xlabel style={font=\color{white!15!black}},
xlabel={$f$},
ymin=0,
ymax=1,
scaled y ticks = false,
ylabel style={font=\color{white!15!black}},
ylabel={TDR},
axis background/.style={fill=white},
legend style={at={(0,1)},anchor=north west,legend cell align=left, align=left, draw=white!15!black},
legend columns=2
]

\addplot[dotted,  color=red, line width=1pt ]
table {%
20	0.183644203206178
40	0.248495028312112
60	0.278376254177139
80	0.295971021660600
100	0.307226137691136
120	0.313533791989673
140	0.320774511033047
160	0.324414805134385
180	0.328380156107222
200	0.333640736713493
220	0.337944298156188
240	0.343868430318459
260	0.348677840355927
280	0.353614654842604
300	0.356833729886285
320	0.360568495153600
340	0.366450306510737
360	0.366912038227153
380	0.367824987581567
400	0.371900426081486
420	0.374226010247978
440	0.377673729848221
460	0.381158147326607
480	0.382229607046962
500	0.389797419161085
520	0.396757765407244
540	0.395627123428351
560	0.393908894572018
580	0.398385584684952
600	0.401619045934054
620	0.405338028649318
640	0.411019371605585
660	0.403185765712384
680	0.404355495425630
700	0.403456305318030
720	0.408978371189666
740	0.410519222593861
760	0.409829239140683
780	0.407215028532024
800	0.410968990250413
820	0.411050235536622
840	0.411233978480382
860	0.408671162119282
880	0.406969136733189
900	0.415345527769944
920	0.419097836674800
940	0.420934519503825
960	0.419827201449255
980	0.427218739419866
1000	0.427297579726793
};
\node[black, font=\footnotesize] at (axis cs:120,0.25) {$L = 25$};

\addplot[dashed, color=blue, line width=1pt, mark size=1pt]
table {%
20	0.335917593536285
40	0.473412251341481
60	0.520516504101500
80	0.543821958657683
100	0.557568676359224
120	0.568772990175662
140	0.573104368168946
160	0.583811232459336
180	0.600667214101190
200	0.606780366403166
220	0.611050310957862
240	0.625799097323729
260	0.631760916001705
280	0.632638960049410
300	0.641447872174927
320	0.649884780007600
340	0.657510547882299
360	0.661705351594238
380	0.670459854975391
400	0.673934247290660
420	0.673684233273364
440	0.675623938553518
460	0.678733781173651
480	0.683428543662790
500	0.689288418015985
520	0.689212206786797
540	0.696766136594312
560	0.698869251456055
580	0.703380925709625
600	0.706861595464270
620	0.709553522464847
640	0.713213975740964
660	0.713559952487105
680	0.715029916179707
700	0.717861415377045
720	0.721880997332364
740	0.724135702123917
760	0.720978736634772
780	0.730986362882365
800	0.733137117327508
820	0.733573912461041
840	0.734519679078624
860	0.743376939059420
880	0.748782752553808
900	0.756185898861618
920	0.760007211487667
940	0.761459368841614
960	0.767325914772967
980	0.769860004387437
1000	0.766586305272536
};
\node[black, font=\footnotesize] at (axis cs:200,0.5) {$L = 50$};

\addplot[ dashdotted, color=orange, line width=1pt, mark size=1pt]
table {%
20	0.401411801121661
40	0.566147775846445
60	0.628017196740095
80	0.661491331771181
100	0.681365767392367
120	0.696496372920391
140	0.707168648222626
160	0.718225284942633
180	0.725458444453334
200	0.724674256908259
220	0.731556906290740
240	0.744686562061461
260	0.754686116056581
280	0.763968546607813
300	0.770461305562495
320	0.781043491595799
340	0.788095812740983
360	0.795314822724024
380	0.801970928184922
400	0.803720188073115
420	0.806147712795923
440	0.811693079422279
460	0.818232207393622
480	0.822275758903678
500	0.828890649139714
520	0.825451071535591
540	0.834569243383590
560	0.835119076719863
580	0.835141674360415
600	0.836496360684301
620	0.835782297719471
640	0.836584460606205
660	0.835839619247524
680	0.840895285996240
700	0.842965243393045
720	0.842248133173486
740	0.846473065168750
760	0.845807991176108
780	0.855009633193951
800	0.858561236183249
820	0.858846228140487
840	0.864368983372625
860	0.868824167961018
880	0.868965864011243
900	0.872467151390714
920	0.878131931114968
940	0.880702996412142
960	0.882441373565513
980	0.882316919092063
1000	0.882712098561270
};
\node[black, font=\footnotesize] at (axis cs:300,0.7) {$L = 75$};

\addplot[ solid, color=cyan, line width=1pt, mark size=1pt]
table {%
20	0.445657110654398
40	0.627466711609551
60	0.705666629025169
80	0.733021757972684
100	0.743750119247950
120	0.755428907535693
140	0.761099429311629
160	0.767461750194471
180	0.778464739528953
200	0.788008935026281
220	0.794702315515022
240	0.803154022197530
260	0.813921691190508
280	0.816913539589400
300	0.824558569199914
320	0.827026197417833
340	0.836534357224430
360	0.841281818134613
380	0.847271235556554
400	0.851523979311835
420	0.856994550274357
440	0.861164189177700
460	0.865563524508828
480	0.869237590065356
500	0.876340018309651
520	0.879241142895966
540	0.884418079270220
560	0.887486605571562
580	0.892739359404340
600	0.896201104903845
620	0.895585278366854
640	0.899027412430629
660	0.899946223959094
680	0.901145308296941
700	0.902805311997184
720	0.903080849369426
740	0.905301664469226
760	0.906300860362030
780	0.911271959055383
800	0.913901290955731
820	0.914539599371054
840	0.916971421136507
860	0.921133100936890
880	0.920458560285701
900	0.920830337492254
920	0.921781985469070
940	0.922330783360762
960	0.925614655328101
980	0.925643285595281
1000	0.923444546396608
};
\node[black, font=\footnotesize] at (axis cs:400,0.9) {$L = 100$};

\end{axis}

\end{tikzpicture}
    \caption{\gls{tdr} versus the time frame $f$ for track-and-stop with different values of the number of uplink slots $L$.}
    \label{fig:sensitibitity_analysis_L}
\end{figure}
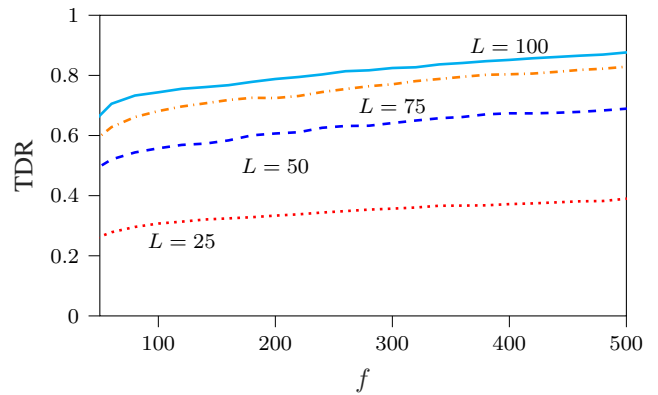

Fig.~\ref{fig:sensitibitity_analysis_L} shows the \gls{tdr} as a function of the frame index $f$ for different numbers of uplink slots $L$. From Fig.~\ref{fig:sensitibitity_analysis_L}, we can see that \gls{tdr} becomes larger as $L$ increases because the reader can execute anomaly detection task based on  more number of received spikes, realizing accurate decision based on the e-value in~\eqref{eq:e_value} and decision threshold defined in \eqref{eq:alpha_update_rule}. For example, to achieve \gls{tdr} larger than 0.9 under $\Delta_{\mathrm{max}} = 0.5$, there is a need for $L = 100$ uplink slots.
 
\subsubsection{Impact of anomaly probability $\pi_1$}
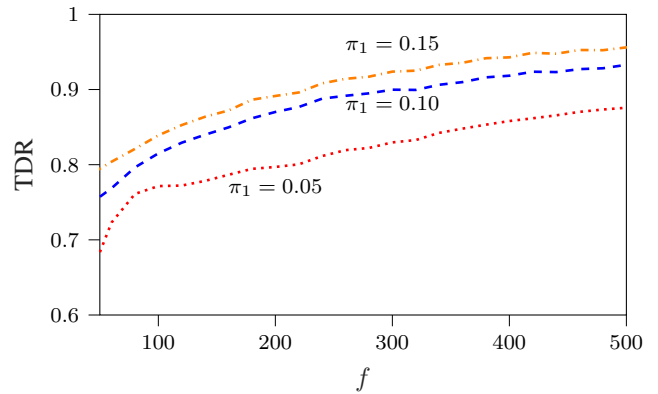
\begin{figure}[t]
    \centering
    \begin{tikzpicture}
\definecolor{crimson2143940}{RGB}{214,39,40}
\definecolor{darkgray176}{RGB}{176,176,176}
\definecolor{darkorange25512714}{RGB}{255,127,14}
\definecolor{forestgreen4416044}{RGB}{44,160,44}
\definecolor{lightgray204}{RGB}{204,204,204}
\definecolor{mediumpurple148103189}{RGB}{148,103,189}

\def\hsep{1.2cm}
\def\vsep{1cm}
\def\vside{2.8cm}
\def\hside{0.18\textwidth}

\begin{axis}[%
width=7cm,
height=4cm,
scale only axis,
xmin=50,
xmax=500,
xlabel style={font=\color{white!15!black}},
xlabel={$f$},
ymin=0.6,
ymax=1,
scaled y ticks = false,
ylabel style={font=\color{white!15!black}},
ylabel={TDR},
axis background/.style={fill=white},
legend style={at={(0.4,0.4)},anchor=north west,legend cell align=left, align=left, draw=white!15!black},
legend columns=2
]

\addplot[dotted,  color=red, line width=1pt ]
table {%
20	0.467872666647174
40	0.643968983520341
60	0.722953035851868
80	0.761111739833948
100	0.771499608536675
120	0.772226888850693
140	0.778461659453128
160	0.786583025394286
180	0.794434129204327
200	0.796994437008340
220	0.800436974053249
240	0.811511506982839
260	0.819356217605511
280	0.822285746789561
300	0.829656342602506
320	0.832769921307311
340	0.842339945191843
360	0.847996417993253
380	0.853242675715242
400	0.858077978128207
420	0.861720727378502
440	0.865653806124902
460	0.870283860578324
480	0.873427420018047
500	0.875780757054106
520	0.875830738068298
540	0.881527872798247
560	0.882664932655579
580	0.887884923669548
600	0.888521365727625
620	0.890487278001027
640	0.897088828561472
660	0.897496446399051
680	0.897941679729612
700	0.897305809689696
720	0.897551003660690
740	0.900425025891541
760	0.900763432559450
780	0.904708316002612
800	0.906714354260785
820	0.910159362577217
840	0.914560017743575
860	0.917442867018743
880	0.917345234036698
900	0.919571862561687
920	0.918607994658881
940	0.915453465485275
960	0.915865527591891
980	0.920563194951125
1000	0.920316760092234
};
\node[black, font=\footnotesize] at (axis cs:200,0.77) {$\pi_1 = 0.05$};

\addplot[dashed, color=blue, line width=1pt, mark size=1pt]
table {%
20	0.643439847937674
40	0.745857099670605
60	0.768662672711261
80	0.796358178844773
100	0.815061476953038
120	0.829266082078119
140	0.839919844786905
160	0.849650072160234
180	0.861739979267507
200	0.870136677837375
220	0.876726328931123
240	0.887859276710393
260	0.891690541052107
280	0.894914117924280
300	0.899783344577048
320	0.899475484149861
340	0.906320917167474
360	0.909891275472345
380	0.916206823378875
400	0.918388988747327
420	0.923687741397556
440	0.923213245895551
460	0.927158622921502
480	0.928286329844822
500	0.932871409346383
520	0.932454140162663
540	0.938420054215339
560	0.937020920556898
580	0.940381682796011
600	0.942724549504247
620	0.942311402814300
640	0.945345990968349
660	0.945214662940599
680	0.948735615404433
700	0.950625871641232
720	0.950215438360623
740	0.953839784845094
760	0.951764632094993
780	0.953132714908910
800	0.955614973138972
820	0.953468664213042
840	0.955534167886974
860	0.957104279948439
880	0.954001394215533
900	0.957835452217530
920	0.961341141276685
940	0.959519645750103
960	0.962394791747254
980	0.964950461652353
1000	0.961706049164940
};
\node[black, font=\footnotesize] at (axis cs:300,0.88) {$\pi_1 = 0.10$};

\addplot[ dashdotted, color=orange, line width=1pt, mark size=1pt]
table {%
20	0.723781944570595
40	0.785323708704368
60	0.803907302895294
80	0.821011653154919
100	0.839015405783066
120	0.852711182472051
140	0.863786664691217
160	0.872135974975259
180	0.886439558002267
200	0.891438892527042
220	0.896064348109036
240	0.908233955297191
260	0.914247330752466
280	0.917090090637338
300	0.923817458044978
320	0.925120963925355
340	0.932829106849048
360	0.935602588020718
380	0.941669666293032
400	0.942823375302132
420	0.948849905164797
440	0.947628660166081
460	0.952596318611517
480	0.952330086481733
500	0.956064727729281
520	0.955375510225493
540	0.959416796427603
560	0.958979816693865
580	0.961995768483816
600	0.965373382545020
620	0.964476336446856
640	0.966417468543456
660	0.963997734668788
680	0.966797615977247
700	0.968684634524801
720	0.965730521345746
740	0.968960416810843
760	0.965467634740705
780	0.967985624261272
800	0.969995345835012
820	0.967891721181169
840	0.970307912334282
860	0.972979281316856
880	0.969965487989678
900	0.972358187711692
920	0.973805761100015
940	0.969587078648143
960	0.971635026335336
980	0.973602690918461
1000	0.970424753471628
};
\node[black, font=\footnotesize] at (axis cs:300,0.96) {$\pi_1 = 0.15$};

\end{axis}

\end{tikzpicture}
     \caption{\gls{tdr} versus the time frame $f$ for track-and-stop scheme with different values of anomaly potion $\pi_1 = \{0.05, 0.10, 0.15\}$ }
    \label{fig:Anomaly_portion_analysis}
\end{figure}

Fig.~\ref{fig:Anomaly_portion_analysis} shows \gls{tdr} against $f$ for track-and-stop, where we set the anomaly probability to $\pi_1 = \{0.05, 0.10, 0.15\}$. As $\pi_1$ increases, the probability that the reader obtains the higher e-value in the single frame also increases, making it more likely that the reader rejects the hypothesis at the current frame. As a result, we can see that \gls{tdr} increases as $\pi_1$ becomes larger.

\subsubsection{On the channel error rate}
\begin{figure}[t]
    \centering
    \begin{tikzpicture}
\definecolor{crimson2143940}{RGB}{214,39,40}
\definecolor{darkgray176}{RGB}{176,176,176}
\definecolor{darkorange25512714}{RGB}{255,127,14}
\definecolor{forestgreen4416044}{RGB}{44,160,44}
\definecolor{lightgray204}{RGB}{204,204,204}
\definecolor{mediumpurple148103189}{RGB}{148,103,189}

\def\hsep{1.2cm}
\def\vsep{1cm}
\def\vside{2.8cm}
\def\hside{0.18\textwidth}

\begin{axis}[%
width=7cm,
height=4cm,
scale only axis,
xmin=0, xmax=0.08,
xtick={0, 0.02, 0.04, 0.06, 0.08},
xticklabels={
\(\displaystyle {0}\),
 \(\displaystyle {0.02}\),
  \(\displaystyle {0.04}\),
  \(\displaystyle {0.06}\),
  \(\displaystyle {0.08}\),
  },
xlabel={$\epsilon$},
xtick style={color=black},
scaled x ticks = false,
ymin=0.4,
ymax=0.8,
scaled y ticks = false,
ylabel style={font=\color{white!15!black}},
ylabel={TDR},
axis background/.style={fill=white},
legend style={at={(0,1)},anchor=north west,legend cell align=left, align=left, draw=white!15!black},
legend columns=2
]

\addplot[dotted,  color=red, line width=1pt ]
table {%
0.0	0.692116983971465
0.02	0.6501863078383573
0.04	0.6157355944237032
0.06	0.5359787118475747
0.08	0.505495026029899

};
\node[black, font=\footnotesize] at (axis cs:0.025,0.75) {$\epsilon_{01} = \epsilon$, $\epsilon_{10}= 0$};

\addplot[dashed, color=blue, line width=1pt, mark size=1pt]
table {%
0.0	0.705455594867993
0.02	0.6945923307378636
0.04	0.6672962636290614
0.06	0.672223992612522
0.08	0.6568858870668094

};
\node[black, font=\footnotesize] at (axis cs:0.065,0.6) {$\epsilon_{01} = 0$, $\epsilon_{10} = \epsilon$};

\addplot[ dashdotted, color=orange, line width=1pt, mark size=1pt]
table {%
0.0	0.6998970232953063
0.02	0.6429182379656179
0.04	0.586857591831806
0.06	0.5276104525638822
0.08	0.44592481435960496

};
\node[black, font=\footnotesize] at (axis cs:0.065,0.42) {$\epsilon_{01} = \epsilon$, $\epsilon_{10} = \epsilon$};

\end{axis}

\end{tikzpicture}
     \caption{\gls{tdr} versus error rate $\epsilon$ for track-and-stop.}
    \label{fig:epsilon_change}
\end{figure}
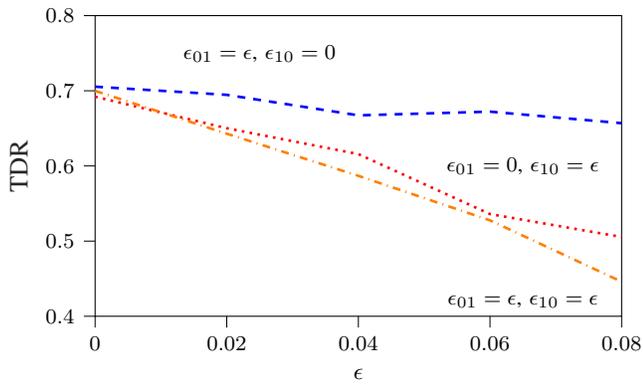

To evaluate the impact of channel error in the binary asymmetric channel described in Sec.~\ref{sec:transmission_model}, we vary the error probability $\epsilon$, by setting: (\emph{i}) $\epsilon_{01} = \epsilon$ and $\epsilon_{10} = 0$; (\emph{ii}) $\epsilon_{10} = \epsilon$ and $\epsilon_{01} = 0$; (\emph{iii}) $\epsilon_{01} = \epsilon_{10} = \epsilon$.   
From Fig.~\ref{fig:epsilon_change}, we see that the error probability $\epsilon_{10}$ has the largest impact on \gls{tdr}. In fact, as increase in $\epsilon_{10}$ causes spike losses, which may be critical for anomaly detection. Conversely, as the error probability $\epsilon_{01}$ increases, it becomes more likely that the reader mistakenly detects false spikes, causing the reader to mistakenly detect an anomaly event when the environment is in a normal state. As the reader must satisfy the \gls{fdr} constraint in \eqref{eq:FDR_condition}, in this case the reader must take a conservative decision when the actual anomaly event happens, reducing the \gls{tdr}. When both errors have a non-zero probability, it is affected by both spike losses and falsely detected spikes, causing further decreases in the \gls{tdr}.

\section{Conclusion} \label{sec:con}
This paper proposed a low-power online anomaly detection framework for neuromorphic wireless sensor networks. 
We have introduced an algorithm that can control the false discovery rate by incorporating  e-values test statistics calculated based on the received spike sequences at each frame. We have also introduced a track-and-stop methods for scheduling nodes based on the channel constraint. Our numerical results show that the proposed approach can track the nodes having the largest spike efficiently with high \gls{tdr}, while maintaining the \gls{fdr} below the predetermined threshold. 

Future work includes designing pull-based communication protocol in neuromorphic communication while taking
into account aspects such
as information freshness and value of information. Designing neuromorphic communication protocols 
for heterogeneous neuro-SNs with different capabilities in terms of sensing and communication is also an interesting
research avenue.

\bibliographystyle{IEEEtran}
\bibliography{IEEEabrv,Ref}

\begin{thebibliography}{10}
\providecommand{\url}[1]{#1}
\csname url@samestyle\endcsname
\providecommand{\newblock}{\relax}
\providecommand{\bibinfo}[2]{#2}
\providecommand{\BIBentrySTDinterwordspacing}{\spaceskip=0pt\relax}
\providecommand{\BIBentryALTinterwordstretchfactor}{4}
\providecommand{\BIBentryALTinterwordspacing}{\spaceskip=\fontdimen2\font plus
\BIBentryALTinterwordstretchfactor\fontdimen3\font minus \fontdimen4\font\relax}
\providecommand{\BIBforeignlanguage}[2]{{%
\expandafter\ifx\csname l@#1\endcsname\relax
\typeout{** WARNING: IEEEtran.bst: No hyphenation pattern has been}%
\typeout{** loaded for the language `#1'. Using the pattern for}%
\typeout{** the default language instead.}%
\else
\language=\csname l@#1\endcsname
\fi
#2}}
\providecommand{\BIBdecl}{\relax}
\BIBdecl

\bibitem{lichtsteiner2006128}
P.~Lichtsteiner, C.~Posch, and T.~Delbruck, ``A 128 x 128 120db 30mw asynchronous vision sensor that responds to relative intensity change,'' in \emph{2006 IEEE Int. Solid State Circuits Conf. Dig. Tech. Papers}, 2006, pp. 2060--2069.

\bibitem{martini2022lossless}
M.~Martini, J.~Adhuran, and N.~Khan, ``Lossless compression of neuromorphic vision sensor data based on point cloud representation,'' \emph{IEEE Access}, vol.~10, pp. 121\,352--121\,364, 2022.

\bibitem{davies2018loihi}
M.~Davies, N.~Srinivasa, T.-H. Lin, G.~Chinya, Y.~Cao, S.~H. Choday, G.~Dimou, P.~Joshi, N.~Imam, S.~Jain \emph{et~al.}, ``Loihi: A neuromorphic manycore processor with on-chip learning,'' \emph{IEEE Micro}, vol.~38, no.~1, pp. 82--99, 2018.

\bibitem{liu2014event}
S.-C. Liu, T.~Delbruck, G.~Indiveri, A.~Whatley, and R.~Douglas, \emph{Event-based neuromorphic systems}.\hskip 1em plus 0.5em minus 0.4em\relax John Wiley \& Sons, 2014.

\bibitem{chen2023neuromorphic}
J.~Chen, N.~Skatchkovsky, and O.~Simeone, ``Neuromorphic wireless cognition: event-driven semantic communications for remote inference,'' \emph{IEEE Trans. Cogn. Commun. Netw.}, vol.~9, no.~2, pp. 252--265, 2023.

\bibitem{skatchkovsky2020end}
N.~Skatchkovsky, H.~Jang, and O.~Simeone, ``End-to-end learning of neuromorphic wireless systems for low-power edge artificial intelligence,'' in \emph{Proc. 2020 54th Asilomar Conf. Signals Syst. Comput.}, 2020, pp. 166--173.

\bibitem{chen2024neuromorphic}
J.~Chen, S.~Park, P.~Popovski, H.~V. Poor, and O.~Simeone, ``Neuromorphic split computing with wake-up radios: Architecture and design via digital twinning,'' \emph{IEEE Trans. Signal Process.}, vol.~72, pp. 4635--4650, 2024.

\bibitem{chen2022neuromorphic}
J.~Chen, N.~Skatchkovsky, and O.~Simeone, ``Neuromorphic integrated sensing and communications,'' \emph{IEEE Wireless Commun. Lett.}, vol.~12, no.~3, pp. 476--480, 2022.

\bibitem{10946192}
D.~Wu, J.~Chen, B.~Rajendran, H.~Vincent~Poor, and O.~Simeone, ``Neuromorphic wireless split computing with multi-level spikes,'' \emph{IEEE Trans. Mach. Learn. Commun. Netw.}, vol.~3, pp. 502--516, 2025.

\bibitem{lee2024asynchronous}
J.~Lee, A.-H. Lee, V.~Leung, F.~Laiwalla, M.~A. Lopez-Gordo, L.~Larson, and A.~Nurmikko, ``An asynchronous wireless network for capturing event-driven data from large populations of autonomous sensors,'' \emph{Nature Electron.}, vol.~7, no.~4, pp. 313--324, 2024.

\bibitem{schaefer2022aegnn}
S.~Schaefer, D.~Gehrig, and D.~Scaramuzza, ``{AEGNN}: Asynchronous event-based graph neural networks,'' in \emph{Proc. IEEE/CVF Conf. Comput. Vision Pattern Recognit.}, 2022, pp. 12\,371--12\,381.

\bibitem{spectrum}
G.~Rak, ``Salt-size sensors mimic the brain: An array of tiny wireless nodes could someday find their way into brain-machine interfaces,'' [Online], \url{https://spectrum.ieee.org/brain-machine-interface-2667619198}.

\bibitem{alkhatib2014review}
A.~A. Alkhatib, ``A review on forest fire detection techniques,'' \emph{Int. J. Distrib. Sensor Netw.}, vol.~10, no.~3, p. 597368, 2014.

\bibitem{rebjock2021online}
Q.~Rebjock, B.~Kurt, T.~Januschowski, and L.~Callot, ``Online false discovery rate control for anomaly detection in time series,'' \emph{Adv. Neural Inf. Process. Syst.}, vol.~34, pp. 26\,487--26\,498, 2021.

\bibitem{ahmad2017unsupervised}
S.~Ahmad, A.~Lavin, S.~Purdy, and Z.~Agha, ``Unsupervised real-time anomaly detection for streaming data,'' \emph{Neurocomputing}, vol. 262, pp. 134--147, 2017.

\bibitem{lavin2015evaluating}
A.~Lavin and S.~Ahmad, ``Evaluating real-time anomaly detection algorithms--the numenta anomaly benchmark,'' in \emph{2015 IEEE 14th int. Conf. Mach. Learn. Appl. (ICMLA)}.\hskip 1em plus 0.5em minus 0.4em\relax IEEE, 2015, pp. 38--44.

\bibitem{javanmard2018online}
A.~Javanmard and A.~Montanari, ``Online rules for control of false discovery rate and false discovery exceedance,'' \emph{Ann. Statist.}, vol.~46, no.~2, pp. 526--554, 2018.

\bibitem{garivier2016optimal}
A.~Garivier and E.~Kaufmann, ``Optimal best arm identification with fixed confidence,'' in \emph{Conf. Learn. Theory}.\hskip 1em plus 0.5em minus 0.4em\relax PMLR, 2016, pp. 998--1027.

\bibitem{event}
Prophesee, ``Event-based vision applications,'' [Online], \url{https://www.prophesee.ai/}.

\bibitem{event2}
IniVation, ``Extreme machine vision,'' [Online], \url{https://inivation.com/}.

\bibitem{event3}
SynSens, ``Speck,'' [Online], \url{https://www.synsense.ai/products/speck-2/}.

\bibitem{ke2024neuromorphic}
Y.~Ke, Z.~Utkovski, M.~Heshmati, O.~Simeone, J.~Dommel, and S.~Stanczak, ``Neuromorphic wireless device-edge co-inference via the directed information bottleneck,'' in \emph{2024 Int. Conf. Neuromorphic Syst. (ICONS)}.\hskip 1em plus 0.5em minus 0.4em\relax IEEE, 2024, pp. 16--23.

\bibitem{kaufmann2016complexity}
E.~Kaufmann, O.~Capp{\'e}, and A.~Garivier, ``On the complexity of best-arm identification in multi-armed bandit models,'' \emph{J. Mach. Learn. Res.}, vol.~17, no.~1, pp. 1--42, 2016.

\bibitem{zhang2023bayesian}
Y.~Zhang, O.~Simeone, S.~T. Jose, L.~Maggi, and A.~Valcarce, ``Bayesian and multi-armed contextual meta-optimization for efficient wireless radio resource management,'' \emph{IEEE Trans. Cogn. Commun. Netw.}, vol.~9, no.~5, pp. 1282--1295, 2023.

\bibitem{ramdas2024hypothesis}
A.~Ramdas and R.~Wang, ``Hypothesis testing with e-values,'' \emph{arXiv preprint arXiv:2410.23614}, 2024.

\bibitem{shiraishi2024coexistence}
J.~Shiraishi, S.~Cavallero, S.~R. Pandey, F.~Saggese, and P.~Popovski, ``Coexistence of push wireless access with pull communication for content-based wake-up radios,'' in \emph{GLOBECOM 2024 - 2024 IEEE Global Commun. Conf.}, 2024, pp. 4836--4841.

\end{thebibliography}
\end{document}